%% file: main.tex
\documentclass[a4paper,onecolumn]{quantumarticle}
\pdfoutput=1

\usepackage[normalem]{ulem}
\usepackage{graphicx}
\usepackage{subcaption}
\usepackage[english]{babel}
\usepackage[T1]{fontenc}
\usepackage[utf8]{inputenc}
\usepackage{tikz}
\usepackage{lipsum}
\usepackage{amsmath}
\usepackage{amsfonts}
\usepackage{amssymb}
\usepackage{float}

\usepackage[backend=bibtex,style=numeric,sorting=none]{biblatex}
\usepackage{braket}

\usepackage{geometry}
\usepackage{hyperref}
\usepackage{xcolor}
\usepackage{booktabs}

\usepackage{siunitx}

\AtBeginDocument{\RenewCommandCopy\qty\SI}
\usepackage[inline]{enumitem}
\usepackage{pdfpages}
\usepackage{multirow}
\usepackage{longtable}
\usepackage{rotating}
\usepackage{array}
\usepackage{noto}
\usepackage{ulem}
\usepackage{DejaVuSans}
\usepackage{multirow}

\begin{document}

\title{QTabGAN: A Hybrid Quantum-Classical GAN for Tabular Data Synthesis}

\author{Subhangi Kumari}
\email{subhangikumari.rs.cse24@itbhu.ac.in}
\affiliation{
Department of Computer Science and Engineering, Indian Institute of Technology (BHU), Varanasi, India
}

\author{Rakesh Achutha}
\email{ra741@cam.ac.uk}
\orcid{0009-0001-9964-0811}
\affiliation{
Department of Applied Mathematics and Theoretical Physics, University of Cambridge,
Cambridge, United Kingdom
}

\author{Vignesh Sivaraman}
\email{vignesh.cse@itbhu.ac.in}
% \orcid{0000-0002-0626-2808}
\affiliation{
Department of Computer Science and Engineering,
Indian Institute of Technology (BHU),
Varanasi, India 
}

\maketitle

\input{abstract.tex}
\input{introduction.tex}
% \textcolor{red}{from here on, we are not required to write too much. we need to write things very precisely and to the point as we dealing with maths and algorithms. so check all the section for redundancy and sentences with just beat around the bush. Next, the result section generally also has such issues. so take a look at it and let me know. if you are unable to figure out the redundancy you can take help of LLMs. But I would not suggest to just run it through them because then you will not learn. Therefore, first try to read the paras a few times and try to make sense out of them. you should be at least be able to indentify the places where the writing is not clear or redundant or flow is not correct etc. Both of you can do this exercise for your own good. Get back to me asap}

\input{Preliminaries.tex}
\input{Model_architecture.tex}

\input{Methodology.tex}

\input{Experimental_results.tex}

\input{conclusion.tex}
\input{limitations.tex}

% \bibliographystyle{quantum}
% \bibliography{Ref}
\printbibliography

\end{document}

%% file: abstract.tex
%!TEX root=./main.tex
\begin{abstract}
\textbf{Synthesizing realistic tabular data is challenging due to heterogeneous feature types and high dimensionality. We introduce QTabGAN, a hybrid quantum–classical generative adversarial framework for tabular data synthesis. QTabGAN is especially designed for settings where real data are scarce or restricted by privacy constraints. The model exploits the expressive power of quantum circuits to learn complex data distributions, which are then mapped to tabular features using classical neural networks. We evaluate QTabGAN on multiple classification and regression datasets and benchmark it against leading state-of-the-art generative models. Experiments show that QTabGAN achieves up to 54.07\% improvement across various classification datasets and evaluation metrics, thus establishing a scalable quantum approach to tabular data synthesis and highlighting its potential for quantum-assisted generative modelling.}
\end{abstract}

%% file: introduction.tex
\section{Introduction}

%\subsection{The Impact of Machine Learning and Generative Models}
The rapid evolution of machine learning has transformed data-driven decision-making across various domains. Its impact is evident in various domains such as finance \cite{huang2020deep}, healthcare \cite{esteva2019guide}, cybersecurity \cite{diro2018distributed}, transportation \cite{azad2024review} \cite{hassan2025application}, retail \cite{wei2020deep} \cite{weber2019domain}, agriculture \cite{knizia2025harnessing}, manufacturing \cite{wang2018deep}, and energy \cite{mocanu2016deep}. 

The growing need for privacy-preserving, scalable, and accessible data resources has made synthetic data generation an important focus in machine learning. Generative Adversarial Networks~\cite{goodfellow2014generative} are one of the widely used data generation techniques. In general, a GAN consists of a generator that generates synthetic data samples and a discriminator that evaluates the fidelity of these samples to the real data distribution. The GAN is trained in an adversarial setting until the discriminator is unable to distinguish between the generated data samples and the real data.
%Generative models, especially Generative Adversarial Networks, have emerged as a prominent area of study within machine learning, demonstrating their ability to produce realistic data samples that mimic distributions of real data. As proposed by \cite{goodfellow2014generative}, GANs consist of a generator architecture that generates synthetic data and a discriminator architecture that checks its authenticity, trained adversarially to achieve an equilibrium where the generator yields samples that are indistinguishable from real data.

%\subsection{Challenges in Tabular Data Synthesis}
%While traditional GANs excel at generating continuous data such as images, synthesizing tabular data, which often includes a mix of numerical and categorical features, remains challenging due to its heterogeneity and high dimensionality.
%The synthesis of tabular data is highly challenging in practice because of the complexity of real-world datasets, which tend to include mixed data types, i.e., numeric and categorical attributes \cite{xu2019modeling}, complex distributions, data sparsity, high dimensionality, and class-imbalance alongside complex inter-feature correlations, making realistic synthetic data generation challenging.
Traditional GANs perform well on continuous data such as images, but generating realistic tabular data remains difficult because real-world datasets often contain mixed discrete and continuous data \cite{xu2019modeling} \cite{kotelnikov2023tabddpm}, complex and sparse distributions, high dimensionality, class imbalance \cite{adiputra2024ctgan}, and intricate inter-feature dependencies. These factors collectively make synthetic tabular data generation significantly more challenging than continuous data synthesis.
%\subsection{Limitations of Classical GANs}
To address these challenges, several models have been proposed. CTGAN \cite{xu2019modeling}, introduces conditional generation and mode-specific normalization to more easily deal with mixed data types. TableGAN \cite{park2018data} is a GAN-based framework built to create synthetic tabular samples that retain the core statistical behaviour of the real data while reducing re-identification risks, thereby enabling safer data sharing. 
CTAB-GAN~\cite{zhao2021ctab} and its improved variant, CTAB- GAN+~\cite{zhao2024ctab}, incorporate dedicated encoders, downstream task–aware objectives, and differentially private training mechanisms to handle skewed feature distributions and class imbalance. CasTGAN~\cite{alshantti2024castgan} further extends the GAN framework through a cascaded architecture that synthesizes features sequentially, thereby improving dependency preservation and the overall validity of generated samples. In addition, domain-specific GANs have been proposed for applications such as healthcare and network security, demonstrating their effectiveness in generating realistic task-oriented tabular data \cite{alqulaity2024enhanced} \cite{zhao2024enhancing}. 
% \textcolor{red}{suddenly do not talk about conditional GAN. What is conditional GAN you have never mentioned it before. so either just say more recent GANs etc like that but do not add new jargons out of nowhere. it will disturb the flow. also cite something from 2025 if you find, else do that in QGAN.}

% \textcolor{red}{i had mentioned the other day that you should cite papers for everything you claim. this para has not even one citation. how will the reviewer now that you are not making up things simply. Site at least 3-5 papers in the following paras}
Despite these advancements, classical GANs often face challenges to effectively model the intricate distributions of complex tabular datasets, thereby limiting their utility in real-world tasks \cite{borisov2022deep}. Quantum Generative Adversarial Networks utilizes the unique capabilities of quantum computing to enable more efficient representation and sampling of high-dimensional probability distributions \cite{nielsen2010quantum} \cite{biamonte2017quantum} \cite{ngo2023survey}. This quantum advantage potentially overcomes the limitations of classical GANs by capturing complex correlations and generating high-fidelity synthetic tabular data\cite{abbas2021power} \cite{schuld2018supervised}, thereby enhancing performance in various domains where classical approaches fall short due to their inability to fully capture the underlying data distribution. This makes Quantum GANs a compelling choice for overcoming the limitations of classical GANs, providing a more powerful framework for generating realistic synthetic data in real-world scenarios.

%\subsection{Quantum Machine Learning as a Solution}

% Quantum computing offers a promising frontier for enhancing machine learning algorithms by leveraging quantum mechanical principles such as superposition, entanglement, and quantum parallelism, potentially providing computational advantages over classical methods. Quantum GANs (QGANs) extend the GAN framework by incorporating variational quantum circuits (VQCs), to model complex probability distributions \cite{zoufal2019quantum}. Hybrid quantum-classical QGANs, combining a quantum generator with a classical discriminator, leverage the strengths of both paradigms. 
Quantum computing provides a promising direction for enhancing machine learning algorithms by exploiting principles such as superposition, entanglement, and quantum parallelism, that can offer computational advantages over classical approaches. Quantum GANs (QGANs) extend the GAN framework by using variational quantum circuits (VQCs) to model complex probability distributions \cite{zoufal2019quantum}. Typically, a QGAN follows a hybrid approach where a quantum generator is paired with a classical discriminator that allows us to leverage the combined strengths of both computational paradigms.

%\subsection{Gaps in Quantum Generative Models}
% \textcolor{red}{This para is not at all clear. what are you trying to say? where are the gaps? you mention few works on different domains and then suddenly start "this gap...". read and rewrite this para}
Although early implementations of Quantum Generative Adversarial Networks (QGANs) have demonstrated strong potential, most existing literature has concentrated primarily on image-based generation, leaving tabular data generation largely unexplored. Recent studies primarily focus on image generation \cite{chang2024latent} \cite{silver2023mosaiq} using datasets such as MNIST and Fashion-MNIST, while a few works explore time-series modelling for tasks like network-traffic anomaly detection\cite{kalfon2024successive}.
% \textcolor{red}{there is inconsistent in the previous sentence. you talk about early implementation is promising but current research is confined. you rather want to say that QGANs have a lot of potential however, most of the literature has only exoplored the images area and tabular GANs are not yet addressed. in time series line, you can mention about market predictions etc if there are some works there}

% \textcolor{red}{the previous line is very crude and informal. you have to make it formal and more polish. polished does not mean you use very fancy fancy words etc, polished means it should sound soft not very harsh or braggy}. 
% This approach relies on restrictive encoding schemes that limit the expressive utilization of quantum states\textcolor{red}{maybe you want to say that they do not use the full expressiveness of quantum states rather than saying limit the expressive utilization. Again, we should not sound very critical as if we are the reviewer of that paper.} In contrast, our model fully leverages quantum capabilities by employing a minimal number of qubits to effectively explore the entire Hilbert space \textcolor{red}{the sentence does not make sense at all. How do you fully leverage quantum capability by reducing qubits? Reducing the number of qubits is the result of full leverage. So, what are you doing differently that allows you to fully leverage it?}
In the context of tabular data, research in quantum generative modelling is still in its early stages, with TabularQGAN \cite{bhardwaj2025tabularqgan} representing one of the first exploratory attempts. While these preliminary efforts demonstrate potential, the existing literature still lacks a comprehensive framework that fully leverages the capabilities of quantum computing. This gap highlights the need for a Quantum GAN capable of capturing complex distributions and generating high-fidelity synthetic samples. To the best of our knowledge, we address this gap by presenting the first framework whose generative model fully utilizes quantum computational capabilities.

% \textcolor{red}{you want to say that in the existing literature has preliminary work on tabular gan but not a comprehensive one that fully leverages quantum capability and captures complex .... and our work is proposed to fill this gap.} \rak{I like this paragraph.}

%\subsection{Proposed Hybrid Quantum-Classical QGAN Framework}
% \textcolor{red}{There are multiple issues here. 1. the flow is not correct. You should talk about all the literature before talking about our contribution. once you start about our contribution then we should not talk about other papers. 2. there is no need to write so much about other paper. you should mention the main idea and the mention the drawback and how we are addressing it. 3. the flow seems to be a bit off in the introduction. Why are we not having a seperate related works section? is this the norm for this journal? Please look into it. If the journal allows a seperate related work section, the we can do it that way. my issue is not to make a seperate section but the flow has to be correct.}

In this paper, we propose a novel Quantum Tabular Generative Adversarial Network (QTabGAN) that generates synthetic tabular data with high fidelity that is statistically representative and preserves label consistency. First, our model uses an $ n$-qubit variational quantum circuit as the generator core to generate \(2^n\) dimensional probability distributions. Next, we deploy a classical neural network to map these distributions to a tabular form. Ultimately, the discriminator serves as the adversary, distinguishing between real and synthetic data samples.
To summarize, the main contributions of our work are as follows.
 \begin{enumerate}
    \item We propose a novel Quantum Tabular GAN framework (QTabGAN) that learns complex data distributions and inter-feature correlations, enabling the generation of high-fidelity, realistic synthetic tabular data.
    \item  Our model leverages the full expressiveness of quantum circuits to enrich the latent representation, enabling the generator to explore complex feature spaces that classical models struggle to capture.
    \item We conduct comprehensive evaluations across diverse real-world tabular datasets using multiple performance metrics, demonstrating that QTabGAN generates high-fidelity synthetic data that remains indistinguishable from real data, highlighting both its robustness and broad applicability across domains.
\end{enumerate}
% \textcolor{red}{the below para adds no value, remove it}
% By achieving these objectives, this work advances quantum machine learning and provides valuable understanding into the practical potential of quantum algorithms for tabular data synthesis.

%\subsection{Organization of the Paper}
Our paper is structured as follows:  Section~\ref{III} presents the preliminaries of quantum computing and provides a high-level description of Generative Adversarial Networks (GANs). Section~\ref{IV} outlines the architecture of the QTabGAN model. Section~\ref{V} details the Experimental Setup, including the data preprocessing strategy, QTabGAN training dynamics and QTabGAN Evaluation. Section~\ref{VI} presents the experimental results, including the dataset description, evaluation metrics, and a comparison of the proposed framework against classical as well as quantum baselines. Section~\ref{VII} discusses the key findings. Section~\ref{Limit} highlights the limitations and identifies opportunities for subsequent research.

%% file: Preliminaries.tex
\section{ PRELIMINARIES}
\label{III}

Quantum computing utilizes the principles of quantum mechanics to process information in ways that are unattainable for classical computers. Quantum bits (qubits) can exist in a superposition of the classical states 0 and 1, allowing them to represent and process multiple states simultaneously. In addition, entanglement creates correlations between qubits, enabling coordinated behaviour across the system. Together, superposition and entanglement enable quantum computers to surpass the computational capabilities of classical machines for certain tasks.
  
\subsection{Qubits and Superposition}
A qubit is the fundamental unit of quantum information, represented as:
\[
|\psi\rangle = \alpha |0\rangle + \beta |1\rangle,
\]
where \(\alpha\) and \(\beta\) are complex probability amplitudes satisfying \(|\alpha|^2 + |\beta|^2 = 1\). The ability of a qubit to exist in both states (i.e. $|0\rangle$ and $|1\rangle$) simultaneously enables quantum algorithms to explore multiple solutions concurrently.

\subsection{Quantum Entanglement}
Entanglement is a quantum phenomenon where the state of one qubit is intrinsically correlated with another, independent of their distance. A maximally entangled two-qubit Bell state is given by
\[
|\Phi^+\rangle = \frac{1}{\sqrt{2}} (|00\rangle + |11\rangle).
\]
Entanglement is an essential resource for quantum computing, enabling complex correlations that classical systems cannot efficiently replicate.

\subsection{Quantum Gates and Circuits}
Quantum gates serve as the fundamental building blocks of quantum circuits. They are unitary operators applied to one or more qubits to perform quantum computations, thereby evolving their states within a complex Hilbert space. Some of the most fundamental gates include:
\begin{itemize}
    \item \textbf{Hadamard Gate (H):} This gate creates an equal superposition of the computational basis states. Its action on the basis state \(|0\rangle\) and \(|1\rangle\) is given by:
    \[
    H|0\rangle = \frac{1}{\sqrt{2}} (|0\rangle + |1\rangle), \quad H|1\rangle = \frac{1}{\sqrt{2}} (|0\rangle - |1\rangle).
    \] 
    \item\textbf{Rotation Gates (Rx, Ry, Rz):} 
    These gates are single-qubit rotation gates that rotate a qubit's state around the x, y, or z-axis of the Bloch sphere by an angle $\theta$.
% \rak{Keep the notation constant every where, sometimes its Rx and sometimes $R_x$ and for R_y and R_z you use exp() notation also but for R_x you don't. Why?}
% Rx gate description and matrix
\paragraph{\(R_x\)($\theta$):} The \(R_x\) gate represents a rotation of a qubit state about the \(x\)-axis of the Bloch sphere by an angle \(\theta\). It is a single-qubit unitary operation that coherently mixes the computational basis states \(\lvert 0 \rangle\) and \(\lvert 1 \rangle\). Its matrix representation is:
\[
R_x(\theta) = \exp\!\left(-i\,\frac{\theta}{2}\,\sigma_X\right)
= \begin{bmatrix}
\cos(\theta/2) & -i \sin(\theta/2) \\
-i \sin(\theta/2) & \cos(\theta/2)
\end{bmatrix}
\]
% Ry gate description and matrix
\paragraph{\(R_y\)($\theta$):}The \(R_y\) gate performs a rotation of a qubit state about the \(y\)-axis of the Bloch sphere by an angle \(\theta\). This single-qubit unitary transformation introduces a coherent mixing of basis states through real-valued amplitudes. Its matrix representation is:
\[
Ry(\theta) = \exp\left(-i \frac{\theta}{2} \sigma_Y\right) = \begin{bmatrix}
\cos(\theta/2) & -\sin(\theta/2) \\
\sin(\theta/2) & \cos(\theta/2)
\end{bmatrix}
\]
% Rz gate description and matrix
\paragraph{\(R_z\)($\theta$):} The \(R_z\) gate corresponds to a rotation about the \(z\)-axis by an angle \(\theta\). Unlike \(R_x\) and \(R_y\) gates, it applies relative phase shifts to the basis states while preserving their amplitudes. Its matrix representation is:
\[
Rz(\theta) = \exp\left(-i \frac{\theta}{2} \sigma_Z\right) =  \begin{bmatrix}
e^{-i\theta/2} & 0 \\
0 & e^{i\theta/2}
\end{bmatrix}
\]
    \item \textbf{Controlled-NOT (CNOT) Gate}: A two-qubit gate that flips the state of the output qubit if the control qubit is in state \(|1\rangle\). It is essential for generating entanglement in quantum circuits:
    \[
    \text{CNOT}(|a\rangle \otimes |b\rangle) = |a\rangle \otimes |a \oplus b\rangle, \quad a, b \in \{0, 1\}.
    \]
\end{itemize}
Quantum circuits are constructed by sequentially applying such gates to initialize, entangle, and transform quantum states in accordance with the computational task.

\subsection{Variational Quantum Circuit}
% \textcolor{red}{VQCs are not hybrid. Make the changes accordingly. Also there is no need to mention about how we are using it here. we will mention that in methodology}
Variational Quantum Circuits (VQCs) are parameterized quantum circuits whose
trainable gate parameters are optimized using classical optimization methods.
They provide a flexible quantum ansatz capable of representing complex quantum
states within a \(2^{n}\)-dimensional Hilbert space. A general VQC acting on \(n\)
qubits can be expressed as

\[
|\psi(\theta)\rangle = U(\theta)\,|0\rangle^{\otimes n},
\]

where \(U(\theta)\) is a unitary operator parameterized by the set of angles
\(\theta\). A common VQC design begins with an initialization layer often using
Hadamard gates to generate superposition followed by layers of parameterized
single-qubit rotations such as \(R_{Y}(\theta)\) and \(R_{Z}(\theta)\).
Entanglement is introduced through controlled operations, most commonly using
CNOT gates:

\[
\text{CNOT}_{i,j}\,|q_i, q_j\rangle = |q_i,\, q_j \oplus q_i\rangle.
\]

Following circuit execution, qubit measurements produce expectation values of
chosen observables, which together define the objective function used during
training. The parameters \(\theta\) are iteratively updated through classical
optimization strategies to refine the circuit’s performance.

\begin{figure*}[t]
    \centering
    \includegraphics[width=\linewidth]{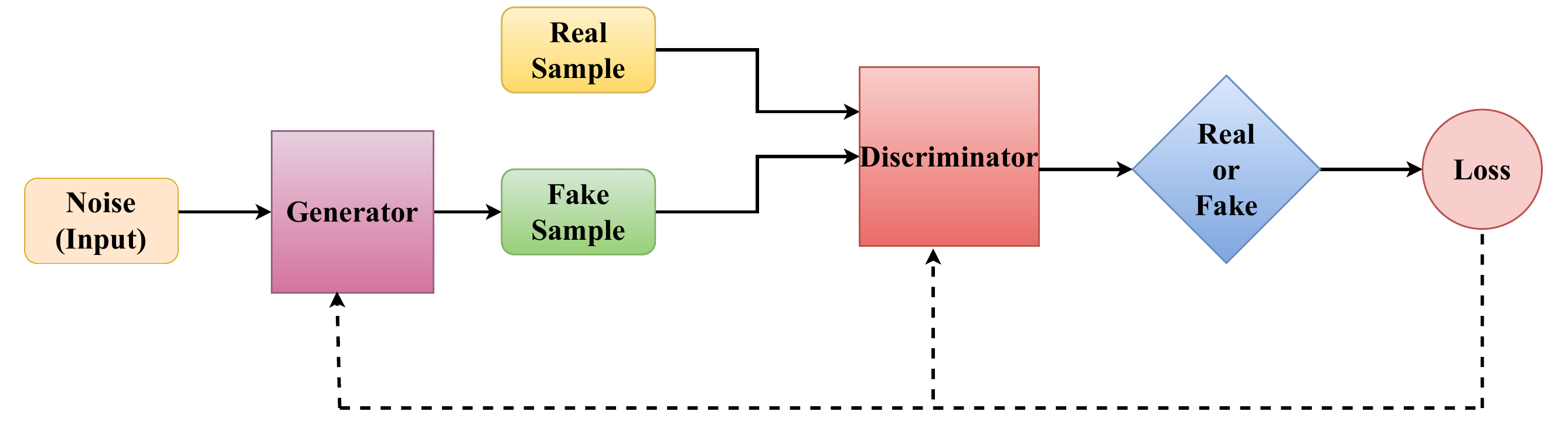}
    \caption{Basic Architecture of Generative Adversarial Networks (GANs)}
    \label{fig:GAN}
\end{figure*}

\subsection{Generative Adversarial Networks (GANs)}
Generative Adversarial Networks are a class of generative models that learn the underlying distribution of a dataset through adversarial training. The basic architecture of Generative Adversarial Networks(GANs) is shown in Figure~\ref{fig:GAN}. A GAN consists of two adversarial components: the generator that maps an input source such as random noise, conditioning information (e.g., class labels), or both together into synthetic data samples, and the discriminator that evaluates whether the given sample is real or synthetic. The neural networks are trained using an adversarial minimax objective where the generator aims to produce samples that the discriminator cannot distinguish, while the discriminator aims to improve its ability to distinguish between the real and synthetic data. This adversarial training enables the generator to gradually synthesize high-fidelity samples that resemble the real-world data. The minimax objective function of a typical GAN is given as follows:
% Generative Adversarial Networks (GANs) are a class of machine learning architectures designed to generate synthetic data similar to a given training set through a competitive process. It consists of two neural networks: a generator and a discriminator. Figure~\ref{fig:GAN} represents the basic architecture of Generative Adversarial Networks(GANs). In this setup, the generator starts with random noise and learns to produce synthetic data samples, whereas the discriminator’s role is to distinguish between the real data taken from the training set and the synthetic data generated by the model. In GAN training, the generator and discriminator are optimized simultaneously in an adversarial minimax framework. The generator incrementally improves its ability to synthesize realistic samples aimed at deceiving the discriminator, while the discriminator is refined to more effectively distinguish real data from generated data. This dynamic interaction is formally expressed through the minimax objective of GANs, defined as:

\begin{equation}
% \begin{split}
\min_G \max_D V(D, G) = \mathbb{E}_{x \sim p_{\text{data}}(x)}[\log D(x)] + \notag \\ \mathbb{E}_{z \sim p_z(z)}[\log (1 - D(G(z)))]
% \end{split}
\end{equation}
Here, $x$ is a real sample and $z$ is a synthetic sample. $D(a)$ is the probability that the discriminator tags a sample $a$ to be real. The discriminator aims to maximize the first term, {\( \mathbb{E}_{x \sim p_{\text{data}}(x)}[\log D(x)] \)}, as the discriminator is supposed to identify the real samples with high probability. In the second term,  \( \mathbb{E}_{z \sim p_{z}(z)}[\log(1 - D(G(z)))]\), $G(z)$ is the sample generated by the generator and $D(G(z))$ is the probability that the discriminator tags the sample to be real. The discriminator aims to minimize this probability, while the generator attempts to maximize it. 

%the first term, \( \mathbb{E}_{x \sim p_{\text{data}}(x)}[\log D(x)] \), encourages the discriminator to
%assign high probability to real samples. The second term,
%\( \mathbb{E}_{z \sim p_{z}(z)}[\log(1 - D(G(z)))] \),
%encourages the discriminator to correctly identify generated samples as fake, while simultaneously guiding the generator to produce samples that the discriminator cannot distinguish from real data.
%The discriminator maximizes this objective to distinguish real from fake samples, while the generator minimizes it to produce data that fools the discriminator. This adversarial optimization process continues until the generator produces data that is indistinguishable from real data, thereby achieving a Nash equilibrium. 

%% file: Model_architecture.tex
\section{Model Architecture}
\label{IV}
\begin{figure*}[htbp]
    \centering
    \includegraphics[width=\linewidth]{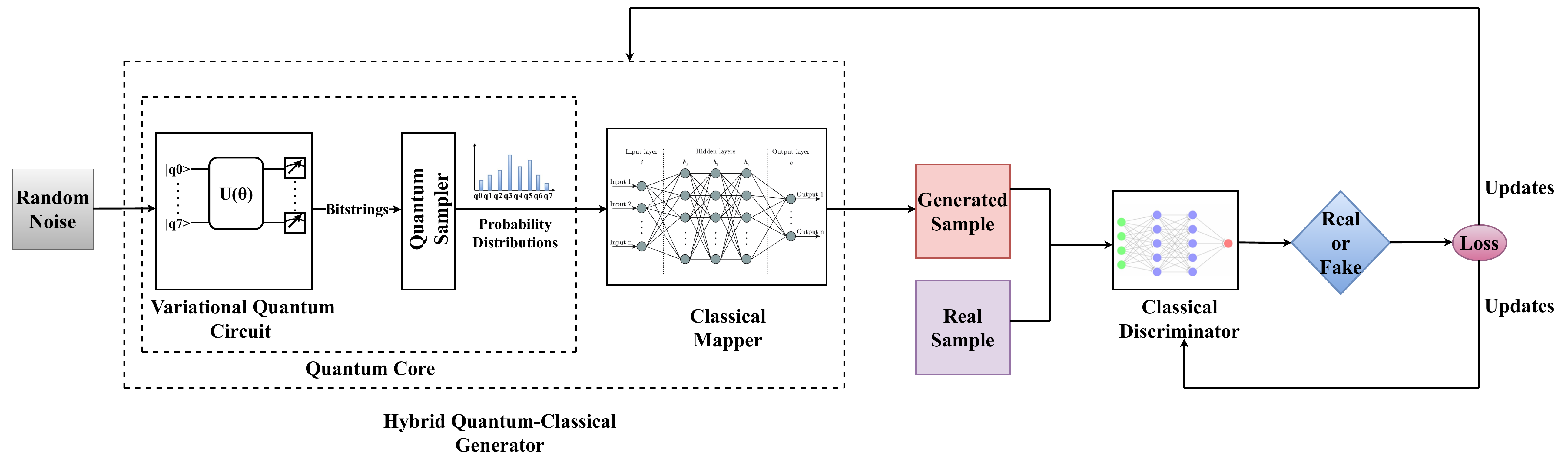}
    \caption{QTabGAN Model Architecture}
    \label{fig:ModelA}
\end{figure*}

Our proposed framework for QTabGAN is a hybrid quantum-classical framework. The model architecture of QTabGAN is shown in Figure~\ref{fig:ModelA}.
The generator's core is a Variational Quantum Circuit (VQC) that generates the probability distributions for synthetic data. VQCs are used to exploit quantum superposition and entanglement. This enables the generator to capture complex data distributions and correlations that classical generators struggle to capture.  Next, the data distributions are mapped to the features of the tabular data using a classical neural network. Finally, the classical discriminator distinguishes between real and synthetic data.

\subsection{Hybrid Quantum-Classical Generator Architecture}
% \textcolor{red}{why are we talking about qskit in the theory part?}
% \textcolor{red}{define the acronym before using it. here Quantum Sampler, what is it?}
Our proposed Hybrid Quantum-Classical Generator is made up of three components, namely, the Variational Quantum Circuit (VQC), the Quantum Sampler, and the Classical Mapper (CLMapper). The VQC and the Quantum Sampler together form the generator's core. The VQC learns the underlying probability distributions of the tabular data and the Quantum Sampler measures the VQC multiple times and gives probability vector. The CLMapper transforms these probability distributions into synthetic data.
% \sout{The generator is a quantum-classical hybrid module made of two components, a \textbf{Quantum Generator} consisting of a Variational Quantum Circuit and a Quantum Sampler, and a \textbf{Classical Mapper (CLMapper)}.}

% \textcolor{red}{remove subsubsection and rather use itemize. the reason here is that by using subsection the font size of the heading has become smaller than the text size which is looking very odd.}\rak{This comment is not yet addressed, yes headings looking smaller than text looks weird.}
\subsubsection{Variational Quantum Circuit} 
\begin{figure*}[htbp]
    \centering
    \includegraphics[scale=0.7]{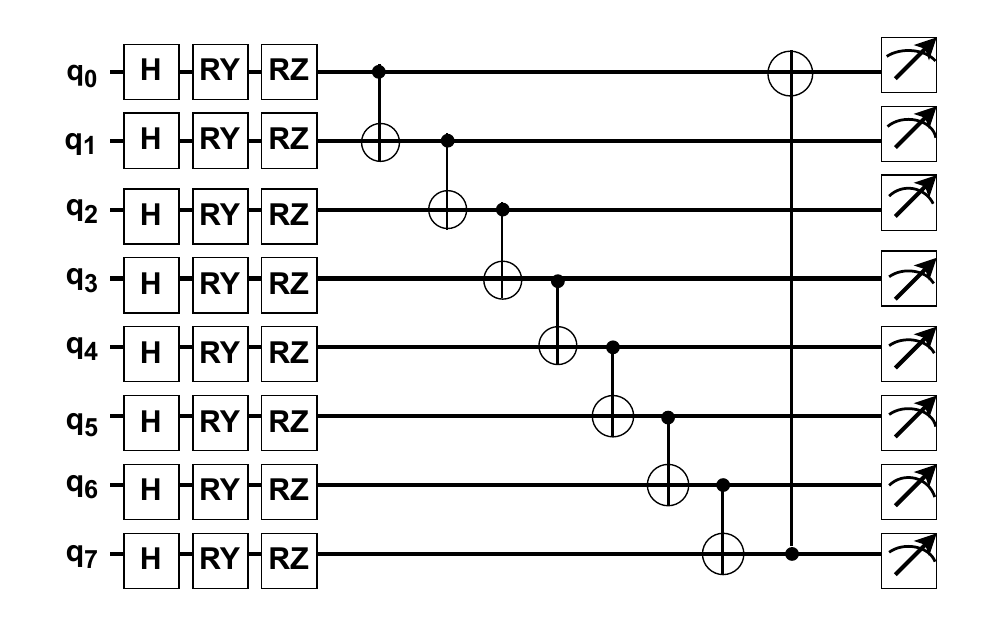}
    \caption{Variational Quantum Circuit Design}
    % \rak{I have told many times till now to keep qubits notation in the figure good and nice, it should look like $q_i$ for all i but in the figure see q1 it looks very odd and improper and also font of q0 is different than others, why?}}
    \label{fig: Quantum Circuit}
\end{figure*}

The Variational Quantum Circuit (VQC) is configured with $n$ qubits. Due to quantum superposition a VQC with $n$ qubits yields a $2^n-$dimensional Hilbert space where each dimension represent a feature of the tabular data. The VQC is constructed with a circuit depth of $L$ layers. Each layer performs state initialization, parameterized rotations, and entanglement operations in the respective order. A single layer of the VQC for $8$ qubits is shown in Figure~\ref{fig: Quantum Circuit}.
% \sout{The Quantum Generator consists of a Variational Quantum Circuit (VQC) to model complex data distributions and a Quantum Sampler to produce an output probability distribution of dimension $2^{n}$. }

% \paragraph{Variational Quantum Circuit Design:}
% Describing the Variational Quantum Classifier configuration
% \textcolor{red}{\sout{it better to write this part in a generic sense. i mean use $n$ for number of qubits instead of 8 and so on. Initially, if you want we can mention we take $n=8$ for our experiments. But never ever mix both, i.e., using somehwere $n$ and somewhere $8$. correct all such issues here.}}
% \sout{The VQC is configured with $n$ qubits, yielding a state space of size $2^{n}$. The circuit is constructed with a depth of $L$ layers, where each layer includes state initialization, parameterized rotations, and entanglement operations. The VQC design is depicted in~ Figure~\ref{fig: Quantum Circuit}.}
% Detailing the initialization process

First, to create a uniform superposition, we initialize each qubit by applying the Hadamard gate. This ensures a robust optimization starting point as all basis states are equally likely. The Hadamard gate is given as follows.
% \sout{Initialization involves applying Hadamard gates to each qubit to create a uniform superposition:}
\[
H^{\otimes n} \ket{0}^{\otimes n} = \frac{1}{\sqrt{2^n}} \sum_{x \in \{0,1\}^n} \ket{x}
\]

% Explaining the parameterized rotations

Next, for every qubit $i$ in every layer $l$, we perform two parameterized rotations as follows.
For each layer $l \in \{1, \cdots, L\}$ and qubit $i \in \{1, \cdots, n\}$, two parameterized rotations are applied as follows:
\[
R_Y(\theta_{i,l}^Y) = \exp\left(-i \frac{\theta_{i,l}^Y}{2} \sigma_Y\right),\ \forall i \in \{1, \cdots, n\}, \forall \ l \in  \{1, \cdots, L\}, (\theta_{i,l}^Y \ | \ \theta_{i,l}^Y \in_{R} [0, 2\pi)
\]
\[
R_Z(\theta_{i,l}^Z) = \exp\left(-i \frac{\theta_{i,l}^Z}{2} \sigma_Z\right)  \forall i \in \{1, \cdots, n\}, \forall \ l \in  \{1, \cdots, L\}, (\theta_{i,l}^Z \ | \ \theta_{i,l}^Z \in_{R} [0, 2\pi).
\]
Here, $\theta_{i,l}^Y$ and $\theta_{i,l}^Z$ are rotation parameters that are chosen randomly in the range $[0,2\pi)$.

Finally, we form a circular entanglement of the qubits using CNOT gates. For every $i \in\{0,(n-2)\}$, we entangle the qubits $i^{\text{th}}$ and $(i+1)^{\text{th}}$ and the $(n-1)^{\text{th}}$ qubit is circularly entangled with the $0^{\text{th}}$ qubit. We formally represent the circular entanglement as follows.
% Describing the entanglement layer
% \sout{We form a circular entanglement by applying CNOT gates between qubits $i$ to $(i+1)\mod n\ \forall 0\leq i \leq n$. We represent the entanglement formally as follows.
% The entanglement layer applies CNOT gates between consecutive qubits ($i$ to $i+1$) and from the last qubit ($q_7$) to the first ($q_0$), forming a circular entanglement or ring-like structure: }
\[
U_{\text{entangle}} = \prod_{i=0}^{n-1} \mathrm{CNOT}\bigl(i, (i+1) \bmod n\bigr)
\]

% \textcolor{red}{i hope the notation of having the qubits in subscript of CNOT is a valid thing}\rak{I agree with this comment, when you are using qubit numbers in subscript explain it in one or two lines after the formula.}

This circular entanglement enhances the VQC's expressive power by
\begin{enumerate*}[label=(\roman*)]
    \item maximizing quantum correlations,
    \item enhancing quantum computational efficiency,
    \item and reducing circuit depth of the VQC~\cite {sim2019expressibility}.
\end{enumerate*}
All three factors together enable the VQC to efficiently capture complex data correlations and distributions. Another advantage of circular entanglement is that it achieves full entanglement across all the qubits with a minimal number of CNOT gates \cite{hein2006entanglement}. This makes it suitable for implementation on Noisy Intermediate-Scale Quantum (NISQ) devices (which are constrained by gate fidelity and coherence times). Once the minimax game of the GAN converges, the statistical properties of the data distribution are encoded in the parameters of the VQC.

\subsubsection{Quantum Sampler}
% \rak{This section is not clear and technically valid, first understand what exactly is quantum sampler, its not vqc based its just to measure the outputs of a VQC it samples from the outputs and give probabilities thats all. No VQC inside it.} 
The quantum sampler is a measurement-based sampling module. This sampler generates samples from the parameterized probability distribution encoded in the quantum state of the VQC. To generate samples, the VQC's quantum state is measured in the computational basis. The VQC is measured $N$ ($N>2^n$) times, and the empirical probability of each possible bitstring is evaluated. This gives us a probability vector of dimension $2^n$. This probability vector is passed to the classical mapper, which maps the probabilities to the different features of the tabular data.

%The next component of the generator's core is the quantum sampler. The sampler executes the VQC. The quantum sampler uses the measurement outcomes that we obtain from VQC and converts them into probability distributions. After the variational quantum circuit is executed, the quantum sampler runs the VQC for a specified number of times to collect measurement outcomes. Running this VQC for specified number of times produces repeated measurements that enable the quantum sampler to estimate the underlying probability distribution. The resulting probability distribution is expressed as a probability vector of dimension $2^{n}$. This probability vector serves as the quantum latent representation that is passed to the classical mapper for further downstream processing.}

\subsubsection{Classical Mapper}
The Classical Mapper (CLMapper) is a classical feedforward neural network that transforms the probability vector ${p}_{\theta}$ into synthetic tabular samples. The generated samples have the same structure as the real dataset.
% The Classical Mapper (CLMapper) is a classical post-processing module that converts the probability distributions obtained from a quantum sampler into a representation suitable for downstream classical learning tasks. In our work, the Classical Mapper (CLMapper) is implemented as a deterministic feedforward neural network that transforms probability vectors produced by a hybrid quantum-classical generator, along with one-hot encoded class labels, into synthetic tabular samples with the same structure and dimensionality as the real dataset.
Let $c$ be the total number of class labels and $d$ be the total number of attributes of the tabular dataset. Let ${y} \in \{0,1\}^c$ denote the one-hot encoded class label of the synthetic data sample to be generated. We can then define the input to CLMapper, ${h_0}$, as the concatenation of ${p_{\theta}}$ and ${y}$. Formally, ${h_0}$ is defined as follows.
\begin{equation*}
\mathbf{h}_0 =
\begin{bmatrix}
\mathbf{p}_{\theta} \\
\mathbf{y}
\end{bmatrix}
\in \mathbb{R}^{2^{n}+c},
\end{equation*}
% \rak{I strongly feel that a diagram of Classical mapper would be very helpful show that we are adding class label at the input and outputs correspond to the tabular attributes. This diagram will be very helpful for reader than the discriminator architecture which is a plain and straight forward neural network.}
The Classical Mapper is implemented as a multi-layer feedforward neural network with $L$ layers whose input is ${h_0}$. Let ${h}_{\ell}$ denote the output of the $\ell$-th layer. For the hidden layers $\ell = 1, \ldots, L-1$, the transformation is defined as
\begin{equation*}
{h}_{\ell} = \sigma\!\left(W_{\ell}{h}_{\ell-1} +{b}_{\ell}\right),
\end{equation*}
where $W_{\ell}$ and ${b}_{\ell}$ are the trainable weight matrix and bias vector of the $\ell$-th layer, respectively, and $\sigma(\cdot)$ denotes the ReLU activation function. The vector ${h}_{\ell}$ represent intermediate feature representations learned by the Classical Mapper.

The final layer applies a linear transformation to produce the output
\begin{equation*}
\hat{\mathbf{x}} = W_{L}\mathbf{h}_{L-1} + \mathbf{b}_{L},
\qquad
\hat{\mathbf{x}} \in \mathbb{R}^{d}.
\end{equation*}
Here, $\hat{\mathbf{x}}$ represents a generated synthetic tabular sample whose dimensionality matches that of the real dataset.

To summarize, the CLMapper takes in $\mathbf{h}_0$ as input and generates the synthetic tabular sample $\hat{\mathbf{x}}$ as the output. The CLMapper is a relation from the Cartesian product of the probability distribution vectors and class labels to the attributes of the tabular dataset. We can formally define this relation as follows.
\begin{equation*}
\mathcal{M}_{\phi} : \mathbb{R}^{2^{n}} \times \{0,1\}^{C} \rightarrow \mathbb{R}^{d}.
\end{equation*}

\subsubsection{Conditional Generation}
% \textcolor{red}{remove redundant definitions and rewrite consicely}
% Conditional generation enhances GANs by
% providing better control over the class of generated data and reduces the chances of the model repeating the same patterns. Along with, probability distributions when class labels are given as input to the Classical Mapper, the generator can produce class-specific samples and still maintain variety across the dataset. This method  helps balance classes in tabular data and also helps to avoid mode collapse, where the generator would output only limited varieties of data.

Conditional generation in our framework enables explicit control over the class of generated tabular samples. This is achieved by concatenating the class label with the quantum-generated probability distribution as input to the Classical Mapper, enabling the generator to produce samples belonging to the desired class. 
% Let $\mathbf{p}_{\theta} \in \mathbb{R}^{2^{n}}$ denote the probability vector obtained by measuring the $n$-qubit quantum circuit, and let $\mathbf{y} \in \{0,1\}^{C}$ be the one-hot encoded class label. The Classical Mapper receives the concatenated input
% \[
% \mathbf{h}_0 = [\mathbf{p}_{\theta}; \mathbf{y}] \in \mathbb{R}^{2^{n}+C},
% \]
% and learns a deterministic mapping
% \[
% \mathcal{M}_{\phi}(\mathbf{p}_{\theta}, \mathbf{y}) = \hat{\mathbf{x}},
% \]
% where $\hat{\mathbf{x}} \in \mathbb{R}^{d}$ is a synthetic tabular sample. 
As a result, the generator learns the conditional distribution $p(x \mid y)$, where $x$ denotes the generated tabular feature vector and $y$ denotes the corresponding class label, rather than the marginal distribution $p(x)$, ensuring that the generated samples are consistent with the provided class label.
This conditioning helps address class imbalance in tabular datasets and reduces the risk of mode collapse, by encouraging the generator to produce more diverse samples. Mode collapse is a issue in generative models where the generator would output only limited varieties of data.  
% \rak{What is mode collapse? Reader doesn't know it automatically.}

The resulting class-conditioned synthetic tabular samples are then forwarded to the discriminator as part of the adversarial training and evaluation pipeline.
\begin{figure*}[htbp]
    \centering
    \includegraphics[scale=0.8]{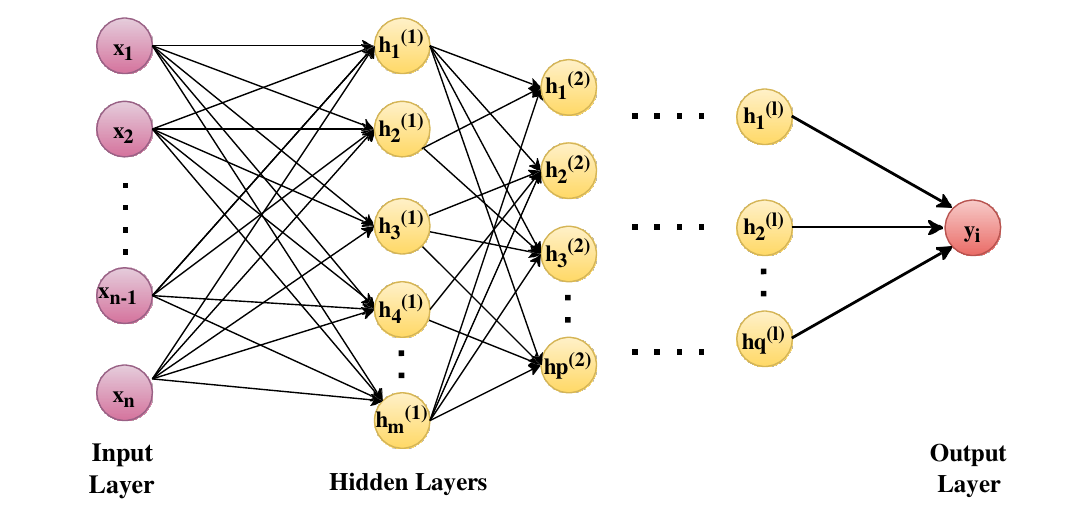}
    \caption{Discriminator Architecture}
    \label{fig: DiscriA}
\end{figure*}
\subsection{Discriminator Architecture}
The discriminator is a densely connected classical feedforward neural network designed to distinguish real tabular samples from synthetic samples generated by the hybrid quantum-classical generator during adversarial training. The discriminator architecture is shown in Figure~\ref{fig: DiscriA}. The discriminator provides feedback to the generator, which guides the generator to produce synthetic samples that closely match the real data distribution.

The discriminator is a multi-layer feedforward neural network with $L$ layers. Let ${h}_{\ell}$ denote the output of the $\ell$-th layer. For the hidden layers $\ell = 1, \ldots, L-1$, the transformation is defined as:
\begin{equation*}
{h}_{\ell} = \sigma\!\left( W_{\ell}{h}_{\ell-1} + {b}_{\ell} \right),
\end{equation*}
where $W_{\ell}$ and ${b}_{\ell}$ are the trainable weight matrix and bias vector of the $\ell$-th layer respectively, and $\sigma(\cdot)$ denotes the ReLU activation function. The vectors ${h}_{\ell}$ represent intermediate feature representations learned by the discriminator.
The final layer applies a linear transformation to produce the discriminator output
\begin{equation*}
D(x, y) = W_{L}{h}_{L-1} + {b}_{L},
\qquad
D(x, y) \in {R}.
\end{equation*}
Here, $D(x, y)$ denotes the scalar discriminator score assigned to the input sample, where $x$ represents the tabular feature vector (real or generated) and $y$ denotes the corresponding class label.
This score is subsequently used to define the adversarial training objective through which the discriminator and the generator are jointly optimized during training.

%% file: Methodology.tex
\section{Experimental Setup}
\label{V}
% \begin{figure*}[htbp]
%     \centering
%     \includegraphics[scale=0.5]{images/workflow.pdf}
%     \caption{Evaluation flow for ML Utility of Classification datasets}
%     \label{fig: MLP}
% \end{figure*}

\subsection{Data Preprocessing}

Data preprocessing is essential before training a model, as it enhances the quality and reliability of the dataset. This stage involves converting raw data into a clean and structured format, addressing issues such as missing values, noise, and inconsistencies. Proper preprocessing enables the model to learn more effectively, enhancing its ability to generalize and perform well on new data. %Here, two types of preprocessing are applied: one for numerical features and another for categorical features. 
The preprocessing of numerical features consists of the following steps.
\begin{enumerate}
\item We clip the feature values to the $1^{\text{st}}$ and $99^{\text{th}}$ percentiles. This minimizes the number of outliers and ensures numerical stability.
\item The second step is to use z-score normalization. We compute the mean $\mu$ and standard deviation $\sigma$ of each feature and scale the feature value $x$ to $x' = \frac{x - \mu}{\sigma + \epsilon}$, where $\epsilon$ is a small constant added to avoid division by zero.
\item Next, we perform min-max scaling that scales these features in the interval $[0, \pi]$. The scaled value $x''$ is evaluted as $x'' = \left[ \frac{x' - \min}{\max - \min + \epsilon} \right] \cdot \pi$, where $\min(x')$ and $\max(x')$ denote the minimum and maximum values of the normalized feature across the training dataset.
\item Finally, we add Gaussian noise $\mathcal{N}(0, 10^{-5})$ to $x''$ to improve gradient stability and to reduce overfitting.
\end{enumerate}

% For \textbf{Numerical feature preprocessing}, numerical features are scaled to the interval $[0, \pi]$. For numerical feature preprocessing, the feature values are clipped to the 1st and 99th percentiles to limit outliers and to ensure numerical stability. Then, z-score normalization is used to compute the mean ($\mu$) and standard deviation ($\sigma$) of each feature, which scales the feature value $x$ to $x' = \frac{x - \mu}{\sigma + \epsilon}$. Next, min-max scaling transforms these features to $[0, \pi]$ using $x'' = \left[ \frac{x' - \min}{\max - \min + \epsilon} \right] \cdot \pi$, and Gaussian noise $\mathcal{N}(0, 10^{-5})$ is added to $x''$ to improve gradient stability and to reduce overfitting.

The preprocessing of categorial features primarily consists of converting the feature into a numerical format using one-hot encoding. In this process, each unique category within a categorical feature is represented as a binary vector. For a feature with multiple unique categories, the encoding creates a vector where the position corresponding to a specific category is marked with a 1, and all other positions are marked with 0s. This transformation is applied to each categorical feature, resulting in a set of binary vectors.

\subsection{QTabGAN Training}

The training process of QTabGAN follows an iterative adversarial learning procedure, which involves a hybrid quantum-classical generator and a classical discriminator. The overall training workflow is summarized as follows:

\begin{enumerate}
    \item \textbf{Model Initialization:}  
    The generator core, the classical mapper (embedded within the generator), and the classical discriminator are initialized. This includes initializing the parameters of the quantum circuit, the weights of the classical mapper, and the weights of the discriminator’s neural network.

    \item \textbf{Quantum Sample Generation:}  
    The generator samples random noise and encodes it into the quantum circuit. The quantum circuit employs parameterized quantum gates, including rotation and entangling gates, to generate a quantum state (in the next step, the quantum sampler measures this quantum state).

    \item \textbf{Probability Distribution Estimation:}  
    The quantum sampler executes the quantum circuit for multiple number of shots and returns a probability vector over its measurement outcomes. This probability vector summarizes the stochastic measurement outcomes drawn from the learned quantum probability distribution of the hybrid quantum–classical generator. 
    % \rak{The outcomes are not random.}

    \item \textbf{Classical Mapping and Conditional Generation:}  
    The generated probability vector is concatenated with the corresponding one-hot-encoded class label and passed as input to the Classical Mapper. The Classical Mapper, implemented as a feedforward neural network, transforms this combined input into a synthetic tabular sample that matches the structure and dimensionality of the real dataset.

    \item \textbf{Discriminator Evaluation:}  
    The discriminator receives both real samples from the dataset and synthetic samples produced by the hybrid quantum-classical generator. It processes these samples through its layers and outputs a score indicating whether each sample is classified as real or synthetic.

    \item \textbf{Loss Computation:}  
    The discriminator loss is computed by comparing its predictions against the true labels (real samples labeled as 1 and synthetic samples labeled as 0) using an appropriate adversarial loss function. The generator loss is derived from the discriminator’s feedback, encouraging the generator to produce samples that are difficult for the discriminator to distinguish from real data.

    \item \textbf{Gradient Computation:}  
    Gradients of the loss functions are computed using the backpropagation algorithm. For the discriminator and the classical mapper, standard backpropagation is applied. For the hybrid quantum-classical generator, gradients with respect to the quantum circuit parameters are computed using hybrid techniques such as the parameter-shift rule, which enables gradient estimation for quantum operations \cite{wierichs2022general}. 
    
    % \rak{cite a paper for parameter shift rule.}

    \item \textbf{Parameter Optimization:}  
    The parameters of both the generator and the discriminator are updated using the computed gradients. The discriminator’s parameters are updated to improve its ability to distinguish real and synthetic samples, while the generator’s quantum and classical parameters are updated to produce more realistic samples.

    \item \textbf{Adversarial Update and Iteration:}  
    The generator and discriminator are updated alternately in an adversarial manner, where the discriminator seeks to minimize its loss, and the generator seeks to maximize the discriminator’s loss. This process is repeated multiple times over several training iterations and epochs.

    \item \textbf{Convergence:}  
    Through repeated adversarial updates, the training process gradually converges toward a Nash equilibrium, where the generated synthetic samples become statistically similar to the real data, and the discriminator can no longer reliably distinguish between them.
\end{enumerate}

\begin{figure*}[htbp]
    \centering
    \includegraphics[scale=0.5]{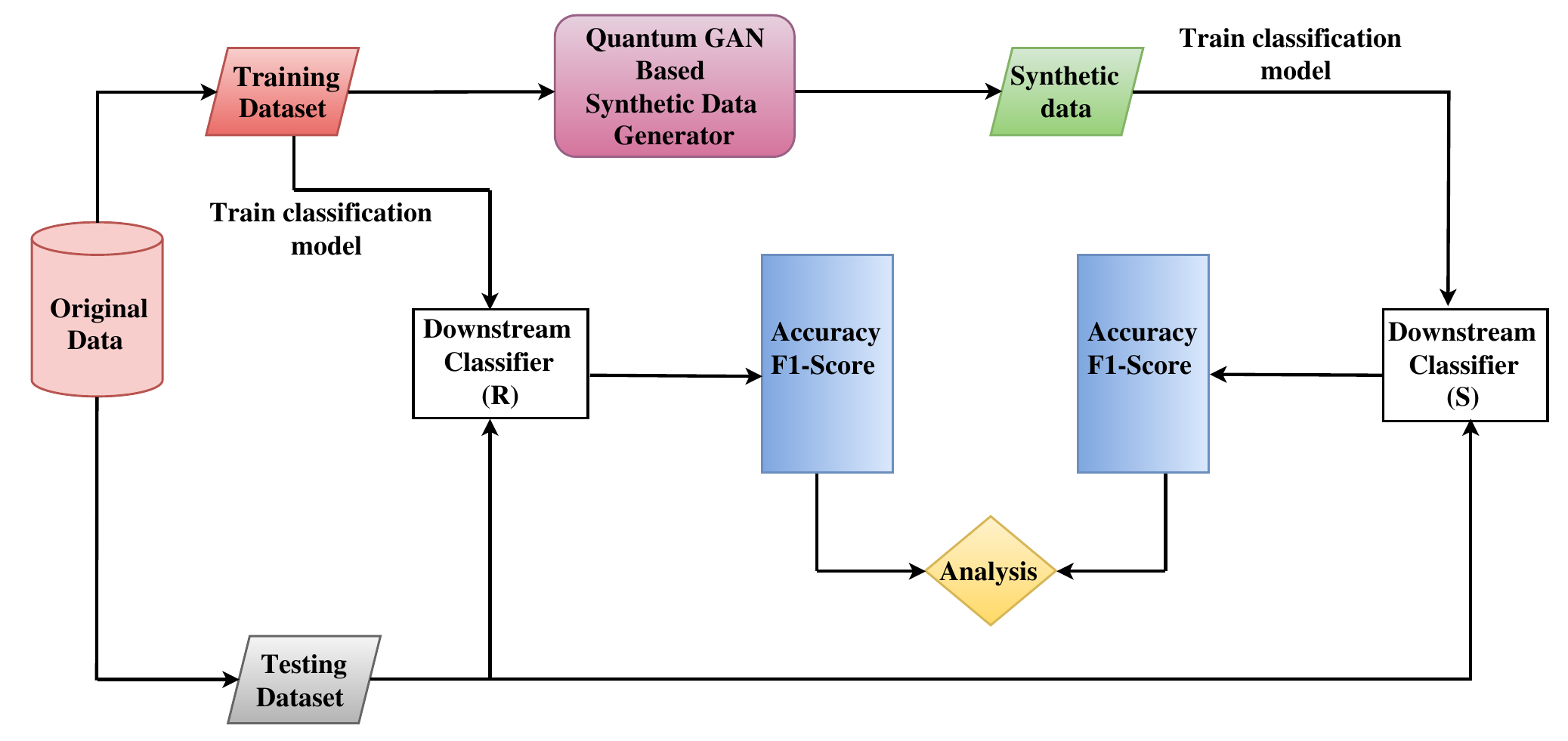}
    \caption{Evaluation flow for ML Utility of Classification datasets}
    \label{fig: MLP}
\end{figure*}

\subsection{QTabGAN Evaluation}
The QTabGAN evaluation framework is shown in Figure~\ref{fig: MLP}. Our evaluation strategy follows established and widely adopted practices in synthetic tabular data generation, as used in prior works such as CTGAN\cite{xu2019modeling}, CTAB-GAN+\cite{zhao2024ctab}, and various other TabularGAN-based models. The framework is designed to assess both the utility and fidelity of the synthetic data produced by QTabGAN. The goal of this evaluation is to assess how well the synthetic data aligns with the statistical properties and predictive performance of the real data when applied in downstream machine learning tasks. 

% \rak{Before proceeding into details we should write that our evaluation strategy is in line with previous standard works, cite some papers and tell that we are using a similar model for evaluating otherwise people may feel out of blue we are giving some new evaluation methods.}

The evaluation workflow is summarized as follows:
\begin{enumerate}
    \item Initially, the original dataset is first divided into two disjoint subsets:
    \begin{enumerate*}[label=(\roman*)]
        \item the training subset and
        \item the testing subset.
    \end{enumerate*}

    \item The training dataset serves two purposes:
    \begin{enumerate*}[label=(\roman*)]
    \item to train QTabGAN that generates synthetic data samples,
    \item to train the downstream classifier (R) for real data.
    \end{enumerate*}
% Initially, the original dataset is divided into two disjoint subsets:
% \begin{enumerate*}[label=(\roman*)]
%     \item the training subset and
%     \item the testing subset.
% \end{enumerate*}
% The training dataset serves two purposes:
% \begin{enumerate*}[label=(\roman*)]
%     \item to train QTabGAN that generates synthetic data samples,
%     \item to train the downstream baseline Multilayer Perceptron classifier for real data (named as MLP(R)).
% \end{enumerate*}
% first, to train QTabGAN, and second, to train a baseline classifier named MLP(R) using real data. 
    \item The QTabGAN is trained repeatedly on the training data until it converges, which helps the model to gradually learn the patterns and structure of the real data distribution. 
    \item Once trained, the model produces such synthetic dataset that replicates the properties of the real data distribution.
% To evaluate the effectiveness of the synthetic data, two separate Multilayer Perceptron (MLP) classifiers are used: (i) trained on real data and (ii) trained on synthetic data. Both classifiers are tested on the same unseen testing dataset to ensure a fair and unbiased comparison.

    \item To evaluate the effectiveness of the synthetic data, two downstream classifiers are used: (i) trained on real data and (ii) trained on synthetic data generated by QTabGAN. Both classifiers are tested on the same unseen testing dataset to ensure a fair and unbiased comparison using metrics such as Accuracy and F1-score.
    
    \item The machine learning utility of the synthetic data is measured by computing the absolute difference between the performance metrics obtained from both the downstream classifiers, given by
    \[
    \left| \text{Metric}_{\text{real}} - \text{Metric}_{\text{synthetic}} \right|.
    \]
    A smaller difference between the performance metrics of the two models indicates that the synthetic data closely mimics the real data and can be effectively used for downstream tasks.

    \item Finally, Statistical similarity is evaluated by comparing the distributions and inter-feature relationships of real and synthetic data, where smaller differences indicate better preservation of real data characteristics.
\end{enumerate}

%% file: Experimental_results.tex
\section{Experimental Analysis}
\label{VI}
In this section, we first compare QTabGAN against state-of-the-art classical data generation methods, followed by a comparison with TabularQGAN, which, to our knowledge, is the only other published work on tabular data generation using quantum GANs. 

% \textcolor{red}{please do not just copy paste lines. For this one I have edited it now.}
%In this section, we evaluate the performance of our proposed QTabGAN against other state-of-the-art classical and quantum data generation techniques. We can broadly categorize our evaluation into two categories: machine learning utility and statistical similarity.  Our results show that QGAN outperforms the other techniques in both categories (and in all the metrics considered).
\subsection{Dataset Description}
To evaluate our QTabGAN, we have used seven different datasets in total. %Out of the seven datasets, five are used for evaluating classification utility, and two are used for evaluating regression utility.
The dataset descriptions are given below in Table~\ref{tab:dataset_description}. The first two datasets are used for regression, while the remaining five are used for classification.
% The description of the datasets used in this work are provided below:\textcolor{red}{give citations for every data set. for credit dataset sydhrClass is binary or numerical}

\subsubsection{King Dataset}
The King dataset contains the records of house sales in King County, Washington. It includes features such as house price, number of bedrooms and bathrooms, square footage of living and lot areas, floors, waterfront status, view, condition, and grade. We utilize `price' as the target variable. The target variable represents the sale price of houses in King County. This dataset offers a rich source for analyzing housing prices in relation to structural, environmental, and geographical factors. 
% To enhance efficiency and performance in synthesizing house prices, the model architecture differs from the general architecture in certain aspects. The classical mapper is simplified to a two-layer neural network without residual connections, as compared to the general model's three-layer mapper with residuals. The discriminator is reduced to four layers compared to the general model's five-layer discriminator with residuals, optimizing training stability for the dataset's size and variability. 

\subsubsection{Insurance Dataset}
The Insurance dataset used in this study contains records related to individual's health and insurance information. It includes features such as age, gender, body mass index (BMI), number of children, smoking status, and residential region. We choose `charges' as the target variable. The target variable represents an individual's total medical expenses. This dataset can be used to examine how demographic and health-related factors influence insurance costs.

\subsubsection{Adult Dataset}
The Adult dataset used contains records of individual's demographic and employment data. It includes features such as age, workclass, education, marital status, occupation, relationship, race, gender, capital gain, capital loss, hours worked per week, native country, and income level. We choose `income' as the target variable. The target variable is binary, indicating whether an individual’s annual income is less than or equal to \$50,000 or greater than \$50,000. This dataset can be used to analyze patterns in income distribution across various socioeconomic factors.%is designed to predict the income class of individuals based on their socioeconomic factors, which makes it suitable for analyzing patterns in income distribution. 
% The model, which is used to generate synthetic probability estimates for these income classes, modifies the generic architecture in certain aspects to enhance efficiency and performance. The classical mapper is simplified to a two-layer neural network with a softmax output and no residual connections, compared to the general model’s three-layer mapper with sigmoid output and residuals, to better suit the binary classification task of income prediction and reduce computational overhead. The discriminator is reduced to four layers compared to the five-layer discriminator in the general model, optimizing training stability for the dataset’s size and feature complexity. 

\subsubsection{Credit Dataset}
The Credit dataset used in this study contains information about credit card transactions. It contains features such as time, amount, and twenty-eight other anonymized variables (V1-V28) derived from principal component analysis.
This dataset aims to predict whether a given transaction is fraudulent. This dataset is generally used to learn patterns associated with fraudulent credit card behavior.
% which aims at predicting a binary class label indicating fraudulent (1) or non-fraudulent (0) transactions. The dataset utilizes `Class' as the target variable. The target variable is binary, indicating whether a transaction is fraudulent (1) or non-fraudulent (0). This dataset is designed for credit risk analysis, enabling the modelling of patterns associated with fraudulent behaviour. 
% The Quantum GAN model, tailored to generate synthetic class probabilities for credit risk, adapts the general model architecture with key modifications to optimize performance for this high-dimensional and imbalanced dataset. The classical mapper in the generator is simplified to a two-layer neural network with a softmax output and no residual connections, as compared to the general model’s three-layer mapper with sigmoid output and residuals, to better suit the binary classification task and reduce computational complexity. The discriminator is reduced to a four-layer architecture as compared to the general model’s five-layer discriminator, ensuring training stability for the imbalanced class distribution. 

\subsubsection{Intrusion Dataset}
This dataset contains network traffic data for intrusion detection.
It contains features such as duration, protocol type, service, flag, source and destination bytes, and various connection metrics, which aims at classifying network activities as normal or specific types of attacks.
The dataset utilizes `attack\_type' as the target variable. The target variable is categorical, representing multiple classes corresponding to normal traffic and different types of network attacks.
This dataset is ideal for analyzing patterns in network security threats due to its high dimensionality and diverse attack types.
% The classical mapper employs a deeper 5-layer feedforward network with batch normalization, LeakyReLU activations, and a softmax output as compared to the general model's simpler 3-layer classical mapper with ReLU and a residual connection, to better capture the complex feature interactions and multi-class classification requirements of the Intrusion dataset. The discriminator uses a 4-layer architecture with batch normalization and LeakyReLU, as compared to the general model's 5-layer discriminator with ReLU and no batch normalization, to improve training stability for the dataset’s imbalanced class distribution.
 
\subsubsection{Loan Dataset}
The Loan dataset contains customer data from a financial institution. It includes features such as age, experience, income, ZIP code, family size, average credit card spending (CCAvg), education level, mortgage, securities account, CD account, online banking, and credit card usage. The dataset utilizes `Personal Loan' as the target variable. The target variable is binary, indicating whether a customer's loan request is approved. This dataset can be used for learning patterns in credit risk analysis.% This dataset is designed to predict whether a customer is likely to accept a personal loan or not, making it suitable for analyzing financial decision-making patterns.

\subsubsection{Covertype Dataset}
The Covertype dataset used in this study contains records of forest cover type data from four wilderness areas in Colorado. It contains features such as elevation, slope, distances to hydrology, roadways, and fire points, hillshade indices, wilderness area, and soil type.
The target is to predict the forest cover type, indicated by the name `Cover\_Type'. There are seven categories of forest cover type. This dataset is designed for classifying forest cover based on environmental variables, making it ideal for analyzing complex ecological patterns.
% To enhance performance on this high-dimensional dataset, the Quantum GAN architecture is modified relative to the general model architecture. The model employs 8 qubits and 8 layers, compared to the general model’s 8 qubits and 6 layers, to increase model capacity for capturing intricate feature interactions in the dataset. The model utilizes 4096 shots instead of 2048 to enhance the precision of quantum measurements, thereby addressing the dataset’s large size and complexity. The classical mapper is expanded to a 4-layer network compared to the general model’s 3-layer mapper, enabling better handling of the dataset’s 54-dimensional input space and multi-class classification task.

% Defining new column type for centering with wrapping
\newcolumntype{C}[1]{>{\centering\arraybackslash}p{#1}}

% Creating a full-width table for two-column layout
\begin{table*}[htbp]
\centering
\setlength{\tabcolsep}{6pt}
\renewcommand{\arraystretch}{1.2}

\begin{tabular}{|p{0.1\textwidth}|p{0.12\textwidth}|p{0.1\textwidth}|p{0.12\textwidth}|p{0.08\textwidth}|p{0.12\textwidth}|p{0.08\textwidth}|}
\hline
\textbf{Dataset} & \textbf{Task} & \textbf{Train/Test Split} & \textbf{Target Variable} & \textbf{Dataset Size} & \textbf{Domain} & \textbf{Feature Count} \\
\hline
King & Regression & 80:20 & Price & 21,600 & Real Estate & 20 \\
\hline
Insurance & Regression & 80:20 & Charges & 1338 & Healthcare and Insurance & 7 \\
\hline
Adult & Classification & 80:20 & Income & 48,000 & Socio-economic & 14 \\
\hline
Credit & Classification & 75:25 & Class & 50,000 & Finance & 31 \\
\hline
Intrusion & Classification & 80:20 & Attack\_Type & 50,000 & Cyber Security & 42 \\
\hline
Loan & Classification & 70:30 & Personal Loan & 5,000 & Finance & 13 \\
\hline
Covertype & Classification & 80:20 & Cover\_Type & 50,000 & Ecology & 54 \\
\hline
\end{tabular}

\caption{Dataset Description.}
\label{tab:dataset_description}
\end{table*}

% To demonstrate the effectiveness of QTabGAN, we have chosen six widely used machine learning datasets and compared it's performance against various state-of-the-art GAN-based tabular data generators. We evaluate QTabGAN's performance by measuring its downstream machine learning utility and its ability to preserve the statistical patterns present in the real data. Furthermore, we extend our analysis by comparing QTabGAN with the recently proposed quantum GAN approach, TabularQGAN, highlighting that our model achieves superior downstream scores.

\subsection{Evaluation Metrics}
We evaluate our proposed QTabGAN on two different evaluation paradigms. The first is the Machine Learning (ML) Utility for evaluating its predictive performance, and the second is Statistical Similarity for evaluating the distributional fidelity of the generated data. The different metrics used for these paradigms are described in the following subsections.
%Our proposed QTabGAN is tested on seven different datasets. Experiments were conducted on five classification datasets: Adult, Loan, Intrusion, Credit, and Covertype and two regression datasets: King and Insurance. The details of each dataset are shown in table~\ref{tab:dataset_description}.
%To evaluate the synthetic data, for both classification and regression datasets, we consider two evaluation paradigms: (i) ML Utility Difference for predictive performance and (ii) Statistical Similarity Difference for distributional fidelity. \textcolor{red}{cite relevant papers for metrics. in consistent writing. you mention two metrics and then ML ultity in itself is having few more metrics. check if this is the norm or else change the writing. maybe you want to say that we evalute on two different paradigms. one is the ML utility and the second is the Data fidelity}

% \textcolor{red}{the discussion on ML utility should all be under that heading. The statistical similarity should not have those discussions. }

\subsubsection{Machine Learning Utility}
The Machine Learning (ML) Utility evaluates how well the synthetic data (generated by QTabGAN and other baseline models) supports downstream predictive tasks.
%compares the performance between models trained on synthetic data and those trained on real data, This paradigm shows 

To test the performance of QTabGAN on classification datasets, we use the following metrics :

\begin{enumerate}
    \item\textbf{Accuracy:} Accuracy is a classification metric that indicates how many predictions made by a model are correct compared to the total number of samples. Higher accuracy means the model predicts class labels more correctly.
    
    % Accuracy is one of the most commonly used metrics for evaluating the classification performance of machine learning models. It is calculated as the ratio of correctly classified instances: comprising true positives (TP) and true negatives (TN) to the total number of instances, which also includes false positives (FP) and false negatives (FN). Essentially, it represents the proportion of correct predictions made by the model across the entire dataset.
    The accuracy is computed as follows.
        \[
        \text{Accuracy} = \frac{\text{TP} + \text{TN}}{\text{TP} + \text{TN} + \text{FP} + \text{FN}}
        \]
    \item\textbf{F1-Score:} The F1-score is a classification metric that measures how well a model predicts class labels by considering both correct and incorrect predictions. It provides a single score that reflects overall classification performance, especially when the dataset contains imbalanced classes. A higher F1-score indicates better model performance.
    %%%
%     The F1-score is a metric used to evaluate the classification performance of machine learning models by combining precision and recall. It is the harmonic mean of precision and recall, providing a balanced measure of both metrics. A high F1-score indicates good performance in both precision and recall, or a high value in one metric that compensates for a lower value in the other, achieving balanced results.
%
% \begin{equation*}
% \text{F1-Score} = 2 \cdot \frac{\text{Precision} \cdot \text{Recall}}{\text{Precision} + \text{Recall}}
% \end{equation*}
%
The F1-score is computed as follows.
\[
\text{F1-Score} = \frac{2 \cdot \text{TP}}{2 \cdot \text{TP} + \text{FP} + \text{FN}}
\]
\end{enumerate}

To test the performance of QTabGAN on regression datasets, we use the following metrics :
\begin{enumerate}
%     \item \textbf{Mean Absolute Error (MAE)}: The Mean Absolute Error (MAE) measures the average magnitude of errors between the predicted and actual values, providing a direct indication of prediction accuracy in regression tasks.

% The formula for MAE is given by:
% \[
% \text{MAE} = \frac{1}{N} \sum_{i=1}^{N} \lvert y_i - \hat{y}_i \rvert
% \]

% where \(y_i\) is the actual value, \(\hat{y}_i\) is the predicted value, and \(n\) is the number of observations.

     \item\textbf{Explained Variance Score (EVS):} The Explained Variance Score (EVS) is a regression metric that measures how well a model explains the variation in the target variable. It demonstrates how well the model’s predictions align with the actual data. A higher EVS value means the model captures more of the underlying variation in the target variable.
    % is a metric used to evaluate the performance of regression models by measuring the proportion of the total variance in the dependent variable that is explained by the model.
    % EVS is especially practical when one wants to assess the extent to which the model grasps the variability of the target variable.
    The EVS is computed as follows.
    % \textcolor{red}{use the same environment for equations until unavoidable. if using equation use that else use the following but keep it consistent}
    \[
    \text{EVS} = 1 - \frac{\text{Var}(x - \hat{x})}{\text{Var}(x)}
    \]
    
    where \(x\) is the actual value, \(\hat{x}\) is the predicted value, and \(\text{Var}\) denotes the variance.
    
    \item\textbf{Coefficient of Determination (\( \mathbf{R}^{\mathbf{2}} \)):} The coefficient of determination ($R^2$) is a frequently used metric used to evaluate regression models. It measures how much of the variation in the target variable is explained by the model’s predictions. Higher $R^2$ values indicate better model performance.
    % One of the most effectively used metrics of evaluating regression models is the Coefficient of Determination,(\( R^2 \)), which provides the proportion of variance in the dependent variable that is explained by the independent variables.
    The $R^2$ is computed as follows.
    \[
    R^2 = 1 - \frac{\sum_{i=1}^{n} (y_i - \hat{y}_i)^2}{\sum_{i=1}^{n} (y_i - \bar{y})^2},
    \]  
where \( y_i \) is the actual value, \( \hat{y}_i \) is the predicted value, \( \bar{y} \) is the mean of the actual values, and \( n \) is the number of observations.
\end{enumerate}
%     \item \textbf{Mean Squared Error (MSE)}: The Mean Squared Error (MSE) measures the average squared difference between the predicted and actual values, giving more weight to larger errors and providing a sensitive measure of regression accuracy.
%     The formula for MSE is given by:
% \[
% \text{MSE} = \frac{1}{N} \sum_{i=1}^{N} (y_i - \hat{y}_i)^2
% \]
% where \(y_i\) is the actual value, \(\hat{y}_i\) is the predicted value, and \(n\) is the number of observations.
After obtaining the above metric values for models trained on synthetic and real data, we compute the ML Utility Difference for the metrics as the absolute difference between the metric values obtained from synthetic and real data, $\text{Metric}_{\text{diff}}= \left|\text{Metric}_{\text{real}} - \text{Metric}_{\text{synthetic}}\right|.$
The aim is to demonstrate the difference in ML utility when a model is trained on synthetic data versus real data. Smaller differences indicate that the synthetic data closely matches the predictive performance of the real data, making it a reliable substitute.

\subsubsection{Statistical Similarity}
The Statistical Similarity compares the distributions of synthetic and real data, assessing fidelity based on statistical properties. We use the following metrics to evaluate QTabGAN's performance.

\begin{enumerate}
    \item\textbf{Jensen-Shannon Divergence (JSD):}
    The Jensen-Shannon Divergence (JSD) is a symmetric measure of similarity between two probability distributions, based on the Kullback-Leibler divergence. It is often used to compare distributions. Specifically, in the case of GANs, it is used to evaluate the quality of generated data. The JSD is computed as follows.
    \[
    \text{JSD}(A || B) = \frac{1}{2} \left( D_{\text{KL}}(A || M) + D_{\text{KL}}(B || M) \right)
    \]

    where \( A \) and \( B \) represent the real and synthetic data distributions, and \( D_{\text{KL}} \) is the Kullback-Leibler divergence. The JSD, which is a symmetric and bounded metric, is calculated as Average JSD (Avg JSD). We compute the Jensen–Shannon Divergence (JSD) for each categorical feature and report the average value across all categorical columns as a single, interpretable score.
    
    % \rak{Avg of what? Explain clearly like average of all samples or average of different datasets I am not clear of this.}

    \item\textbf{Correlation Difference(Diff. Corr.):}
    The correlation difference metric measures how well the synthetic data preserves the pair-wise linear relationships between features present in the real dataset. It quantifies the average absolute deviation between the correlation matrices of the real and synthetic data. Smaller values indicate better structural similarity.
    The formula for Correlation Difference is given by:
    \[
    \text{CorrDiff} = \frac{1}{K} \sum_{(i,j)} 
    \left| \rho^{\text{real}}_{ij} - \rho^{\text{syn}}_{ij} \right|
    \]
    where, $K$ denotes the total number of unique feature pairs $(i,j)$,
    $\rho^{\text{real}}_{ij}$ denotes the Pearson correlation coefficient between features $i$ and $j$ in the real dataset,
    $\rho^{\text{syn}}_{ij}$ denotes the Pearson correlation coefficient between features $i$ and $j$ in the synthetic dataset,
    $\left| \rho^{\text{real}}_{ij} - \rho^{\text{syn}}_{ij} \right|$ represents the absolute difference in correlation for the feature pair $(i,j)$.
\end{enumerate}

\subsection{Result Analysis}
\label{results}
\begin{figure*}[htbp]
    \centering
    \scalebox{1.0}{ % scale entire figure
    \begin{minipage}{\textwidth}
        % First row
        \begin{subfigure}[b]{0.48\textwidth}
            \centering
            \includegraphics[width=\linewidth]{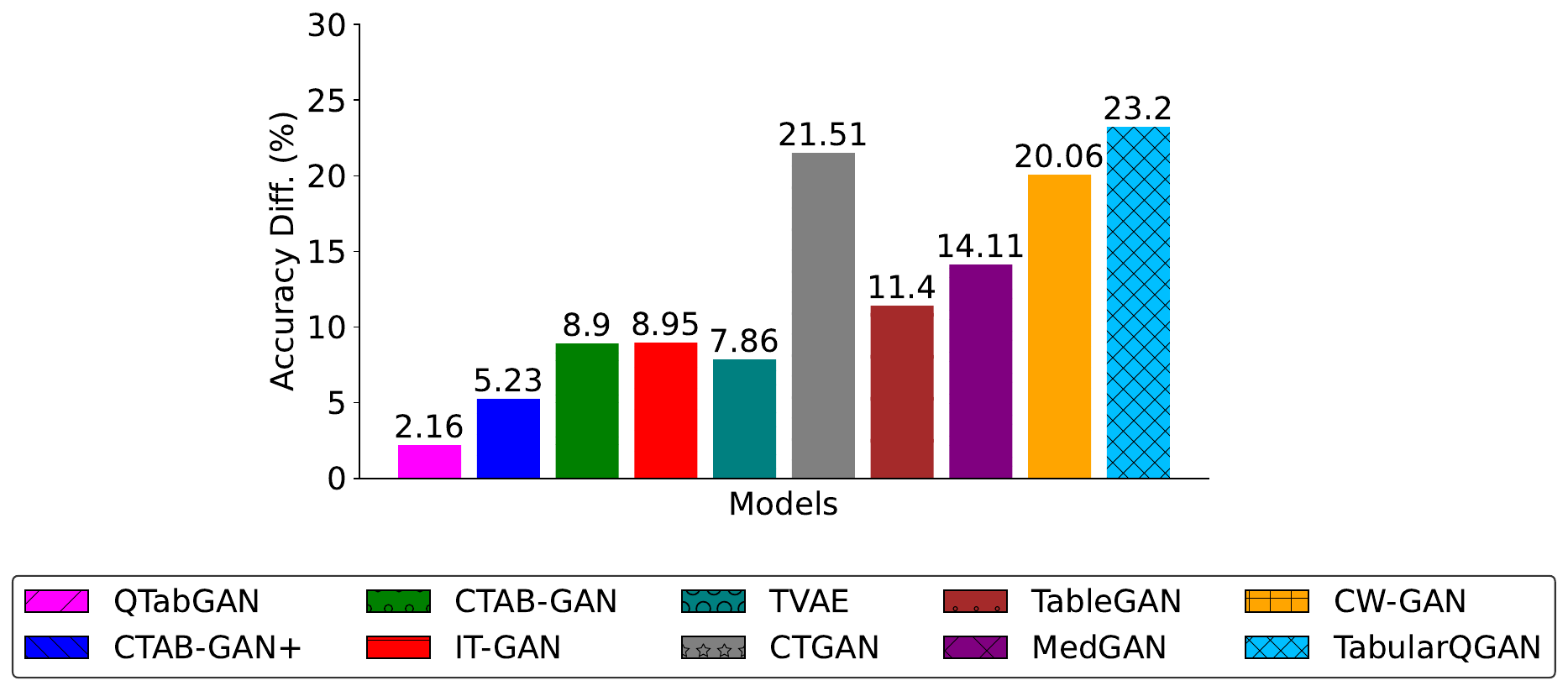}
            \caption{Accuracy Difference}
            \label{fig:subfig1}
        \end{subfigure}%
        \hfill
        \begin{subfigure}[b]{0.48\textwidth}
            \centering
            \includegraphics[width=\linewidth]{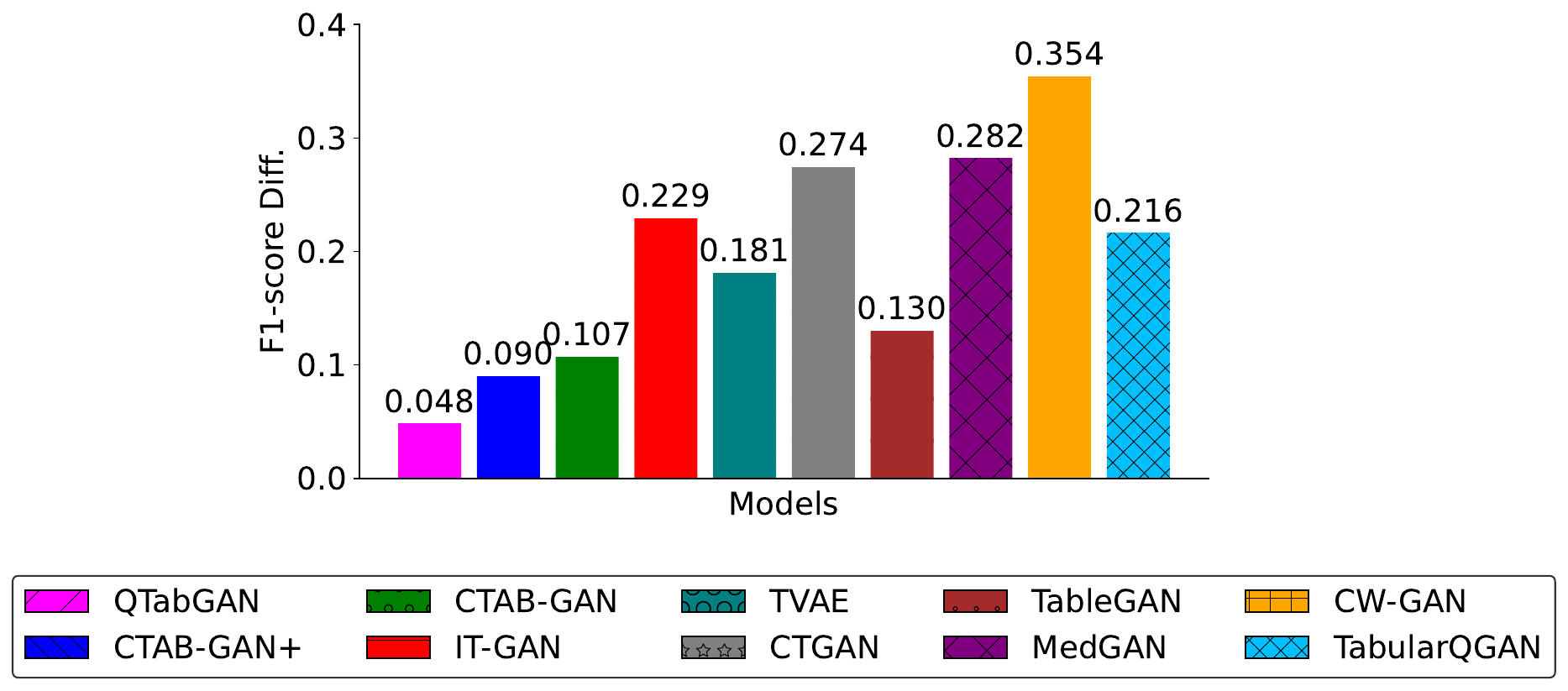}
            \caption{F1-score Difference}
            \label{fig:subfig2}
        \end{subfigure}

        \vspace{0.5cm} % optional space between rows

        % Second row
        \begin{subfigure}[b]{0.48\textwidth}
            \centering
            \includegraphics[width=\linewidth]{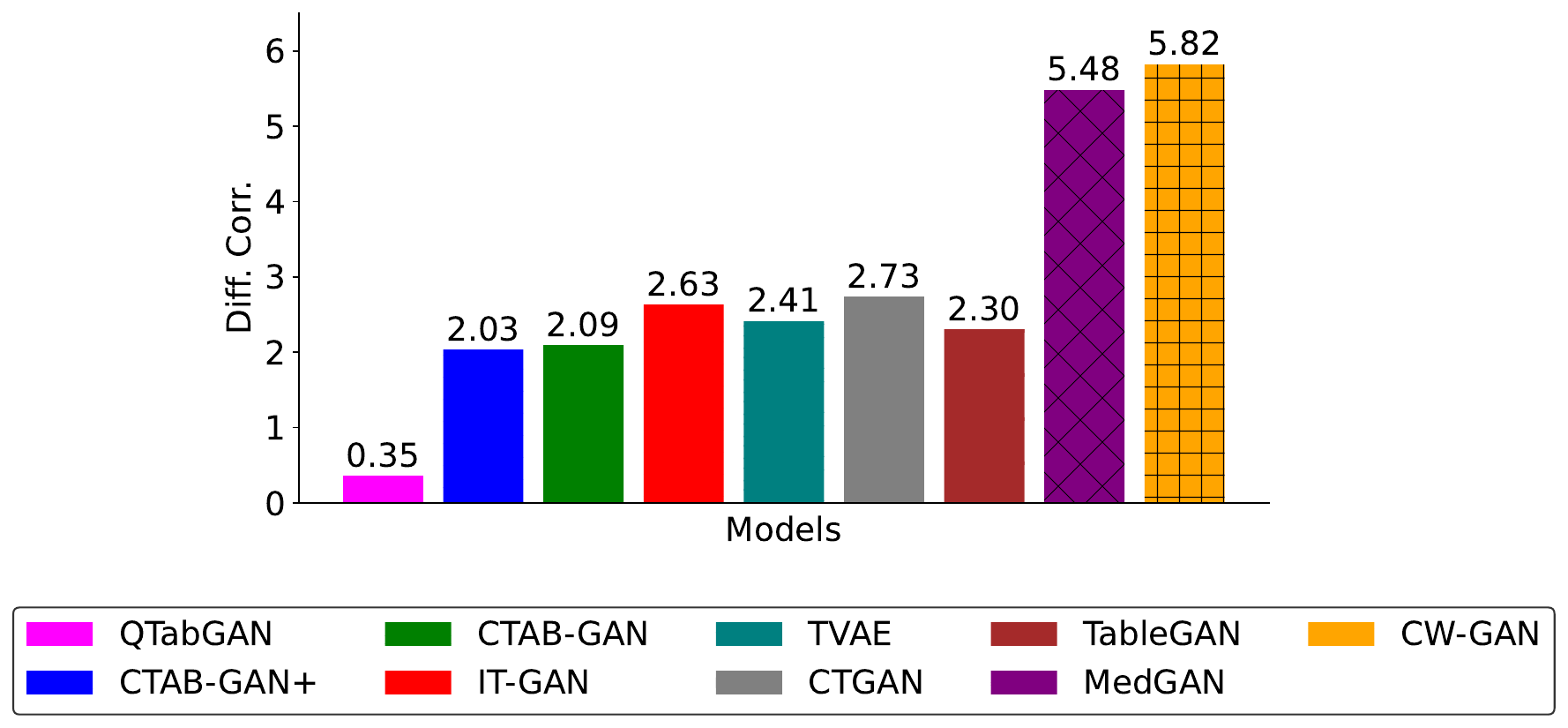}
            \caption{Correlation Difference}
            \label{fig:subfig3}
        \end{subfigure}%
        \hfill
        \begin{subfigure}[b]{0.48\textwidth}
            \centering
            \includegraphics[width=\linewidth]{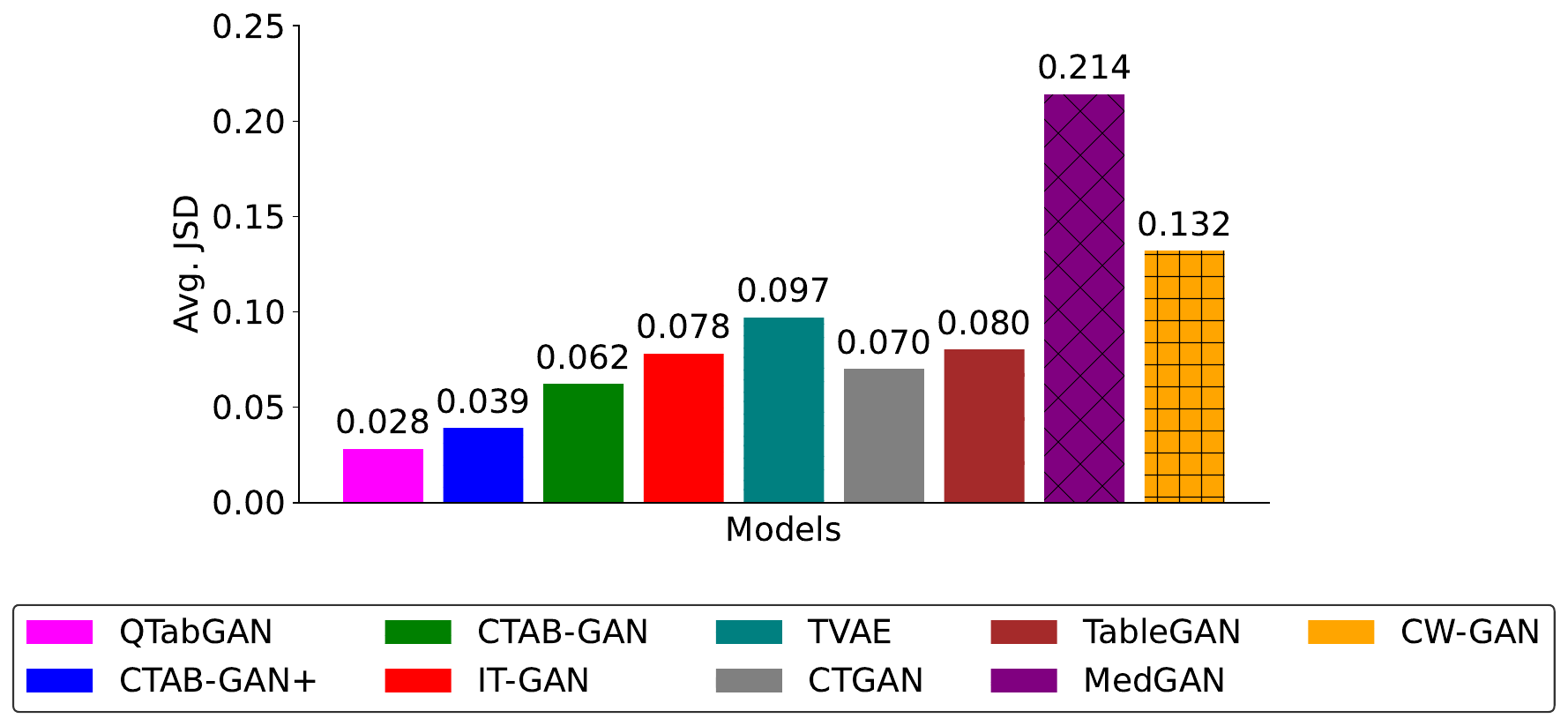}
            \caption{Average JSD}
            \label{fig:subfig4}
        \end{subfigure}
    \end{minipage}
    } % end scalebox

    \caption{Difference in ML utility and statistical similarity between real and synthetic data, averaged across five classification datasets}
    \label{fig:all_subfigures}
\end{figure*}

% To evaluate the performance of our QTabGAN, we conducted experiments on five classification datasets and two regression datasets. The classification datasets include Adult, Loan, Intrusion, Credit and Covertype datasets and the regression datasets include King and Insurance datasets. All the datasets have varying sizes, different feature counts, and different class distributions as shown in table~\ref{tab:dataset_description}.

Next, we discuss the results obtained from our experiments on both classification and regression datasets. For both classification and regression datasets, we compare QTabGAN against eight state-of-the-art generative adversarial network (GAN) models, namely, CTAB-GAN+, CTAB-GAN, IT-GAN, TVAE, CTGAN, TableGAN, MedGAN, and CW-GAN, to evaluate their relative performance across multiple metrics. Among these baselines, CTAB-GAN+ \cite{zhao2024ctab} is used as the main state-of-the-art baseline for comparison. We also compare QTabGAN against the TabularQGAN  \cite{bhardwaj2025tabularqgan} model, which, to our knowledge, is the only other published work on tabular data generation using quantum GANs. For both classification and regression datasets, the results for each metric are obtained by averaging the corresponding metric values across all datasets. The experimental results are illustrated in Figure~\ref{fig:all_subfigures} and Figure~\ref{fig:all_subfigures.} for the classification and the regression datasets, respectively. We first discuss the results on ML utility, followed by an analysis of statistical similarity.

\subsubsection{ ML Utility}

The bar graph presented in Figure~\ref{fig:subfig1} illustrates the differences in accuracy (expressed in percentage). QTabGAN exhibits the lowest accuracy difference of 2.16\%, indicating a relatively minor deviation in its performance. Compared to CTAB-GAN+ (5.23\%) and CTAB-GAN (8.9\%), QTabGAN demonstrates significant improvements, performing approximately 58.7\% better than CTAB-GAN+ and 75.7\% better than CTAB-GAN, reflecting a significantly lower accuracy difference. As compared to TabularQGAN, QTabGAN exhibits an accuracy difference of 2.16\%, whereas TabularQGAN shows a substantially higher difference of 23.2\%. QTabGAN demonstrates significant improvements, performing approximately 90.7\% better than TabularQGAN. This demonstrates that our model significantly outperforms both classical as well as existing quantum baseline in downstream performance.
% \textcolor{red}{can we show f1 score difference also in percentage?}

The bar graph depicted in Figure~\ref{fig:subfig2} presents the differences in F1-scores (expressed as proportions). QTabGAN stands out with the smallest F1-score difference of 0.048, showing a highly consistent performance with minimal variation. In comparison to CTAB-GAN+ (0.090) and CTAB-GAN (0.107), QTabGAN shows remarkable enhancement, achieving approximately 46.7\% superior performance over CTAB-GAN+ and 55.1\% better performance over CTAB-GAN, indicating a significantly reduced F1-score difference. As compared to TabularQGAN, QTabGAN shows a smaller F1-score difference of 0.048, while TabularQGAN shows a F1-score difference of 0.216. This represents a 77.8\% improvement in QTabGAN's performance as compared to the TabularQGAN model. 
% \textcolor{red}{remove the metric and baseline discussion from here. i guess the baselines are same. in regression we are comparing with ctabgan? if yes where is its discussion.}

% Figure~\ref{fig:all_subfigures.} presents a comparative analysis of the ML utility metrics for regression datasets.
% For regression datasets, ML utility is evaluated using metrics such as EVS Difference and R² Difference. These metrics measure how well models trained on synthetic data match the predictive performance of models trained on real data. Lower differences across these metrics indicate higher fidelity and better utility of the synthetic data.
% As shown in Figure~\ref{fig:all_subfigures.}, for regression datasets we compare seven generative adversarial network (GAN) models: QTabGAN, IT-GAN, IT-GAN(Q), VeeGAN, TableGAN, TGAN, and TVAE to evaluate their relative performance across different metrics following the comparison framework presented in \cite{lee2021invertible}.

The bar graph presented in Figure~\ref{fig:subfig1.} illustrates the absolute EVS Difference values. QTabGAN achieves the lowest EVS Difference of 0.02. In contrast, CTABGAN+ and CTABGAN exhibit higher EVS differences of 0.03 and 0.05, respectively. QTabGAN shows a remarkable improvement, achieving approximately 33\% superior performance over CTABGAN+ and approximately 60\% better performance over CTABGAN, showing that QTabGAN matches real-data prediction behaviour more closely. As compared to TabularQGAN, QTabGAN achieves a much lower EVS difference of 0.02, compared to 0.16 for TabularQGAN. This corresponds to an 87.5\% improvement in QTabGAN's performance compared to TabularQGAN. 

% In comparison to CTABGAN+ and CTABGAN, QTabGAN shows significant improvements. It reduces the MAE deviation by 66.67\% compared to IT-GAN and by 50\% compared to IT-GAN(Q), showing that QTabGAN matches real-data prediction behavior more closely.

The bar graph presented in Figure~\ref{fig:subfig2.} illustrates the absolute \(R^{2}\) Difference values. QTabGAN achieves the smallest \(R^{2}\) Difference of 0.02. Compared to CTABGAN+ and CTABGAN, which exhibit higher \(R^{2}\) difference values of 0.04 and 0.06, respectively, QTabGAN achieves a remarkable improvement of approximately 50\% over CTABGAN+ and approximately 66.7\% over CTABGAN, indicating that QTabGAN better captures the variance of the real-data regression model. As compared to the TabularQGAN model, QTabGAN shows a smaller \(R^{2}\) difference of 0.02, whereas TabularQGAN exhibits a substantially higher difference of 3.85. This shows an approximately 99.5\% improvement in QTabGAN's performance compared to TabularQGAN. 

We observe that QTabGAN outperforms both classical GANs and TabularQGAN on downstream ML utility metrics. This behaviour is expected, as QTabGAN efficiently explores a much larger feature space than classical GANs do, thanks to VQC's inherent computational capabilities. Moreover, QTabGAN's better performance compared to TabularQGAN is attributed to their architectural differences. For a given number of qubits (say, ten qubits), TabularQGAN can model only a small number of features (typically three to four), whereas QTabGAN can model all the features by leveraging the classical mapper. This limitation directly affects TabularQGAN's downstream performance, as evidenced by consistently lower ML utility metrics across both classification and regression datasets. Furthermore, we can also observe that TabularQGAN does not even outperform classical models across the ML utility metrics.

\subsubsection{Statistical Similarity}
% Statistical Similarity metrics for classification datasets include Average JSD and Correlation Difference. Figures~\ref{fig:subfig3} and Figure~\ref{fig:subfig4} show the results for the correlation difference and the average JSD metric which are used to evaluate the statistical similarity of the classification datasets. As shown in Figure~\ref{fig:all_subfigures}, for classification datasets we compare seven generative adversarial network (GAN) models: QTabGAN, CTAB-GAN+, CTAB-GAN, CTGAN, TableGAN, MedGAN, and CW-GAN to evaluate their relative performance across multiple metrics, following the comparison framework presented in \cite{zhao2024ctab}.
% \textcolor{red}{I guess we need to modify the way we have added graphs. you can add the ML utility graphs first and then add the stat similarity ones. that way it will be easy for review to see and refer the graphs. also one more thing is that you need to ensure that the figures and tables are present just before their discussion or else on the top of the page where discussion is done. they should not be placed very far. sometimes latex will put it far. check online how to work around that.}

The bar graph presented in Figure~\ref{fig:subfig3} presents the Correlation Difference scores for the classification datasets.
QTabGAN achieves the lowest correlation difference of 0.35, indicating a stronger preservation of inter-feature relationships compared to the baseline models. This improved correlation preservation can be attributed to the use of entangling quantum circuits, which enhance the model’s representational capacity for capturing feature dependencies. In comparison, CTAB-GAN+ and CTAB-GAN show higher deviations of 2.03 and 2.09, respectively. Compared to these two models, QTabGAN outperforms CTAB-GAN+ by approximately 82.76\% and CTAB-GAN by approximately 83.25\%.

The bar graph presented in Figure~\ref{fig:subfig4} presents the average Jensen–Shannon Divergence (Avg. JSD). QTabGAN achieves the lowest Avg. JSD of 0.028. In comparison, CTAB-GAN+ and CTAB-GAN show Avg. JSD scores of 0.039 and 0.062, respectively. Compared to these baselines, QTabGAN achieves significant improvements, reducing divergence by approximately 28.21\% compared to CTAB-GAN+ and by 54.84\% compared to CTAB-GAN, which demonstrates QTabGAN's superior ability to preserve the underlying data distribution.

\begin{figure*}[bthp]
\centering
\begin{subfigure}{0.45\textwidth}
\centering
\includegraphics[width = 1.0 \textwidth]{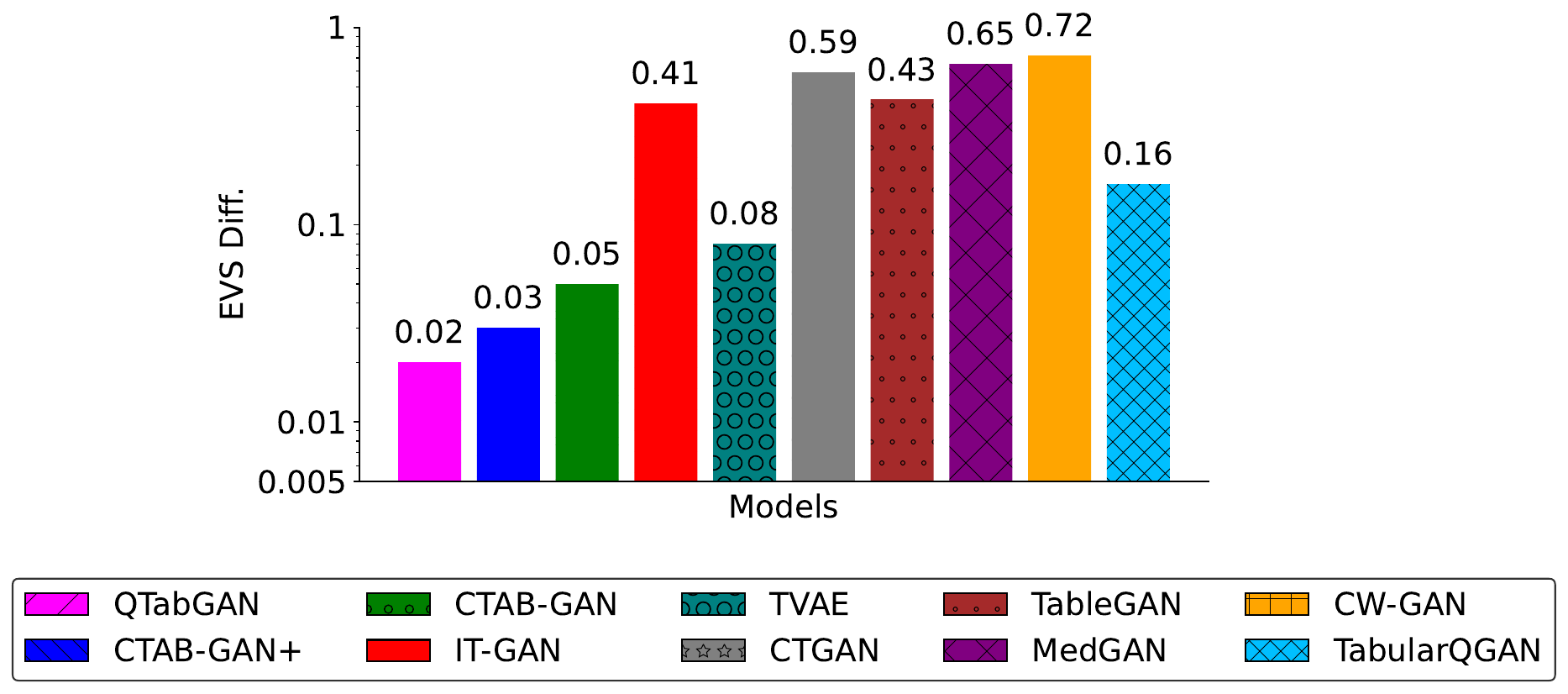}
\caption{EVS Difference}
\label{fig:subfig1.}
\end{subfigure}
\begin{subfigure}{0.45\textwidth}
\centering
\includegraphics[width =  1.0 \textwidth]{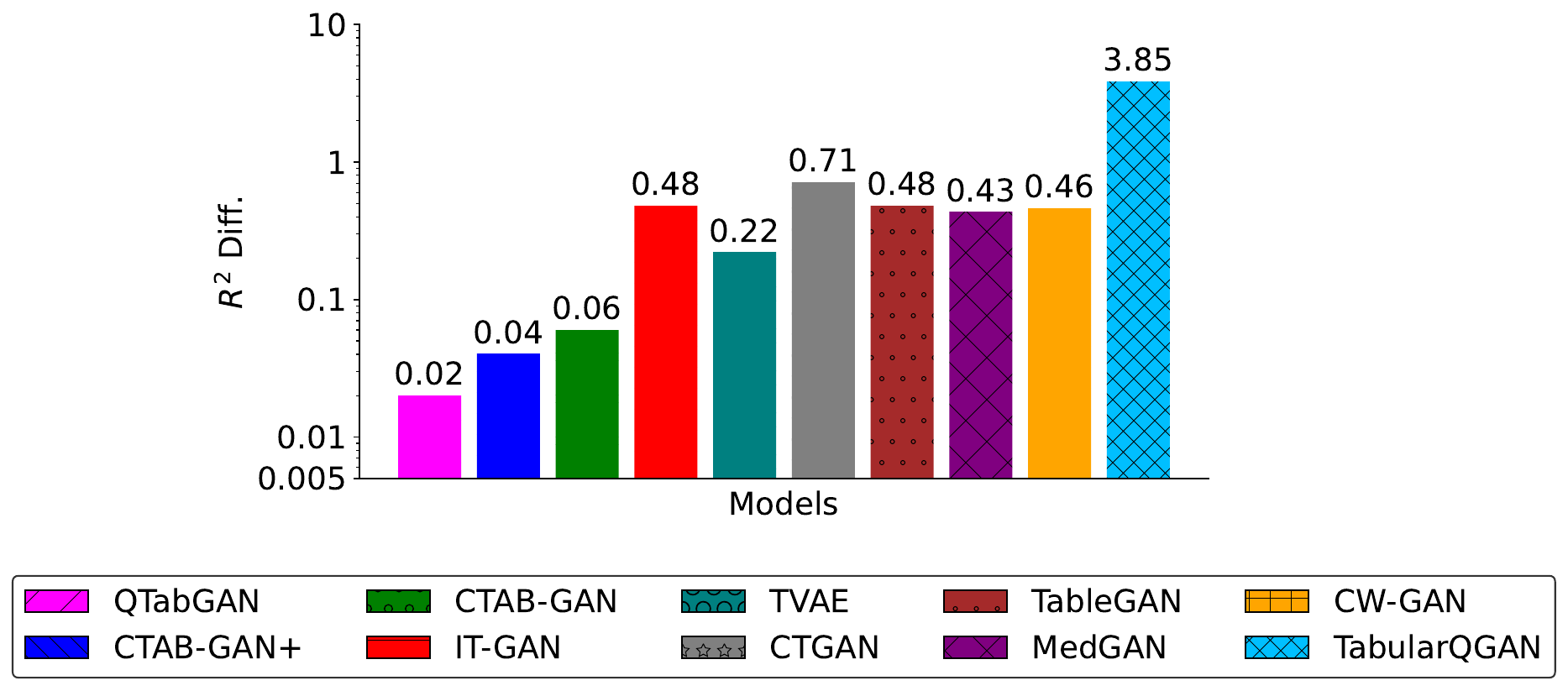}
\caption{$R^2$ Difference}
\label{fig:subfig2.}
\end{subfigure}
\begin{subfigure}{0.45\textwidth}
\centering
\includegraphics[width = 1.0 \textwidth]{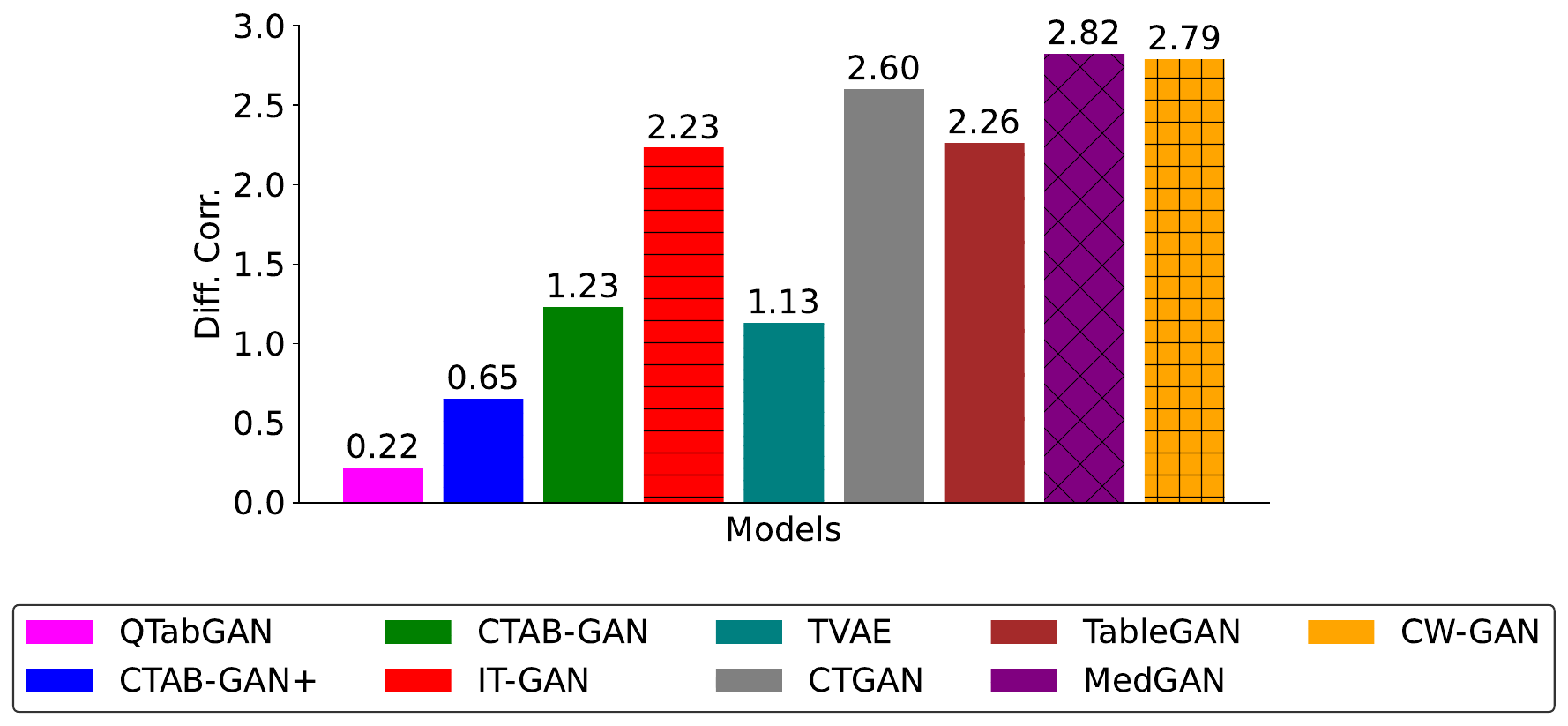}
\caption{Correlation Difference}
\label{fig:subfig3.}
\end{subfigure}
\caption{Difference in ML utility and statistical similarity between real and synthetic data, averaged across two regression datasets}
\label{fig:all_subfigures.}
\end{figure*}

The bar graph presented in Figure~\ref{fig:subfig3.} presents the Correlation Difference scores for the regression datasets.
QTabGAN achieves the lowest correlation difference of 0.22, indicating that it effectively preserves the relationships between features. In comparison, CTAB-GAN+ and CTAB-GAN show higher deviations of 0.65 and 1.23, respectively. Compared to these two models, QTabGAN outperforms CTAB-GAN+ by approximately 66.15\% and CTAB-GAN by approximately 82.11\%. 

We also compare QTabGAN with the TabularQGAN model using statistical similarity metrics across both classification and regression datasets. For the statistical similarity metrics evaluation, both QTabGAN and TabularQGAN are assessed using the same subset of features to ensure a fair and consistent comparison, since these metrics are inherently feature-dependent. The comparative results for both classification and regression datasets are summarized in Table~\ref {tab:comparison_metrics}. Lower values indicate better alignment with the real data distributions.

For classification datasets, QTabGAN achieves a much lower average JSD of 0.05 compared to 0.20 for TabularQGAN. This corresponds to a 75\% reduction in the difference between real and synthetic data distributions, indicating that QTabGAN more accurately captures the marginal feature distributions of the real data.

For classification datasets, QTabGAN achieves a lower correlation difference of 0.40 compared to 0.60 for TabularQGAN, resulting in an approximate 33.3\% improvement in correlation preservation. This indicates that QTabGAN better preserves feature relationships, which helps the synthetic data to more accurately reflect the interactions present in the original dataset.

For regression datasets, QTabGAN achieves a lower correlation difference of 0.14 compared to 0.19 for TabularQGAN, resulting in an approximate 26.3\% improvement in correlation preservation. This shows that QTabGAN more accurately preserves relationships among continuous features, which is important for regression tasks where feature dependencies strongly influence predictive performance.

From the above results for different metrics, it can be observed that QTabGAN outperforms all other state-of-the-art methods in all statistical similarity metrics for both classification and regression datasets.

QTabGAN achieves this strong performance on both classification and regression datasets due to its quantum-based design, which employs a variational quantum circuit to generate expressive probability representations. The hybrid architecture combines a variational quantum circuit with a classical mapping network, enabling the generation of full-feature tabular data in a scalable manner. These design choices allow QTabGAN to produce high-quality synthetic data that closely matches real-data behaviour, thereby reducing performance differences in downstream classification and regression tasks and demonstrating its effectiveness for tabular data generation.

\begin{table}[ht]
\centering
\begin{tabular}{|l|cc|c|}
\hline
Method 
& \multicolumn{2}{c|}{Classification Datasets} 
& \multicolumn{1}{c|}{Regression Datasets} \\
\cline{2-4}
& Avg. JSD & Diff. Corr. 
& Diff. Corr. \\
\hline
QTabGAN        & 0.05 & 0.4  & 0.14 \\
\hline
TabularQGAN   & 0.20  & 0.6 & 0.19 \\
\hline
\end{tabular}
\caption{Comparison of Statistical Similarity Differences for classification and regression datasets between original and synthetic data.}
\label{tab:comparison_metrics}
\end{table}

\subsubsection{Best and Worst Case Analysis}

To better understand the model’s behaviour, we analyze two representative classification datasets: the best-performing dataset (Adult) and one of the worst-performing datasets (Loan). This comparison helps us better understand how QTabGAN behaves in both favourable and more complex scenarios. The detailed accuracy and F1-score results obtained using real and synthetic data for these datasets are summarized in Table~\ref{tab:best_worst_results}. For the Adult dataset, the differences in Accuracy and F1-score between models trained on real and synthetic data are the smallest among all classification datasets considered, indicating that QTabGAN’s synthetic samples closely match the underlying distribution of the real data.

In contrast, the Loan dataset exhibits relatively larger differences in both Accuracy and F1-score. This can be attributed to the smaller sample size and more complex feature relationships present in the Loan dataset. Despite its complexity, QTabGAN performs well on the Loan dataset. These results demonstrate that our model performs well across various types of datasets, ranging from simpler ones, such as the Adult dataset, to more complex ones, like the Loan dataset.

\begin{table}[htbp]
\centering
\renewcommand{\arraystretch}{1.3}   % increase row height
\setlength{\tabcolsep}{8pt}         % column spacing

\begin{tabular}{|l|ccc|ccc|}
\hline
\multirow{2}{*}{\textbf{Metric}} 
& \multicolumn{3}{c|}{\textbf{Best}} 
& \multicolumn{3}{c|}{\textbf{Worst}} \\ \cline{2-7}

& \textbf{Real} & \textbf{Synthetic} & \textbf{Difference}
& \textbf{Real} & \textbf{Synthetic} & \textbf{Difference} \\
\hline
Accuracy & 0.7760 & 0.7740 & 0.0020 & 0.9600 & 0.9380 & 0.0220 \\
F1-score & 0.7619 & 0.7539 & 0.0080 & 0.9570 & 0.9291 & 0.0279 \\
\hline
\end{tabular}

\caption{Best and Worst ML Utility Performance.}
\label{tab:best_worst_results}
\end{table}

% Further, Figure~\ref{fig:best and struggling subfigures} represents the values of metrics such as accuracy and F1-score for real vs synthetic data for best and for one of the challenging datasets. 
% Figure~\ref{fig:best metrics} compares the performance of the classification (Adult) dataset when models are trained on real data versus synthetic data generated by our QTabGAN. For the Adult dataset, the accuracy and F1-score differences between real and synthetic data are the smallest among all the classification datasets used. This shows that QTabGAN’s synthetic samples closely match the real data’s distribution. One possible reason for this strong performance is that the Adult dataset has a relatively large sample size and well-structured categorical and numerical features, which makes it easier for QTabGAN to learn compared to smaller or noisier datasets.
% Figure~\ref{fig:struggling metrics} compares the performance of the classification (Loan) dataset when models are trained on real data versus synthetic data generated by our QTabGAN. The Loan dataset exhibits relatively larger differences in accuracy and F1-score compared to the Adult dataset. This is likely because the Loan dataset has fewer samples and more complex feature relationships. QTabGAN performs well on the Loan dataset despite its complexity. These results demonstrate that our model performs well across various types of datasets, ranging from simpler ones, such as Adult dataset, to more complex ones, like Loan dataset.

\subsubsection{CDF-Based Distribution Analysis}

% Figure ~\ref{fig:King and loan CDF} and Figure ~\ref{fig:credit and adult CDF} shows the smoothed cumulative distribution function (CDF) curves of real and synthetic samples for some representative features of some datasets. A CDF curve is a simple way to describe a data distribution. Smoothed CDFs give a clearer and more stable picture of how closely the synthetic data matches the real data’s overall statistical behaviour.

% For clarity, we present CDF curves for a subset of features for some datasets as examples. These features were chosen to reflect different distributional patterns, such as variation in spread, skewness, and shape, so that the plots illustrate both common and more challenging cases. By doing so, we can demonstrate not only how well the model reproduces simple and regular patterns, but also how effectively it captures more complex or irregular behaviours present in real-world datasets. The same overall behaviour was observed for other features as well, confirming that the similarity between real and synthetic CDFs is consistent across the datasets. This strongly indicates that the generative model preserves the statistical essence of the real data across multiple dimensions, validating its reliability for downstream machine learning tasks and practical applications.

\begin{figure}[H]
    \centering

    % -------- Legend at top-right --------
    \begin{minipage}{\linewidth}
        \hfill
        \includegraphics[width=0.15\linewidth]{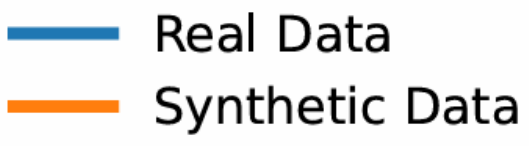}
    \end{minipage}

    \vspace{0.1cm}

    % -------- First row --------
    \begin{subfigure}[t]{0.4\linewidth}
        \centering
        \includegraphics[width=\linewidth]{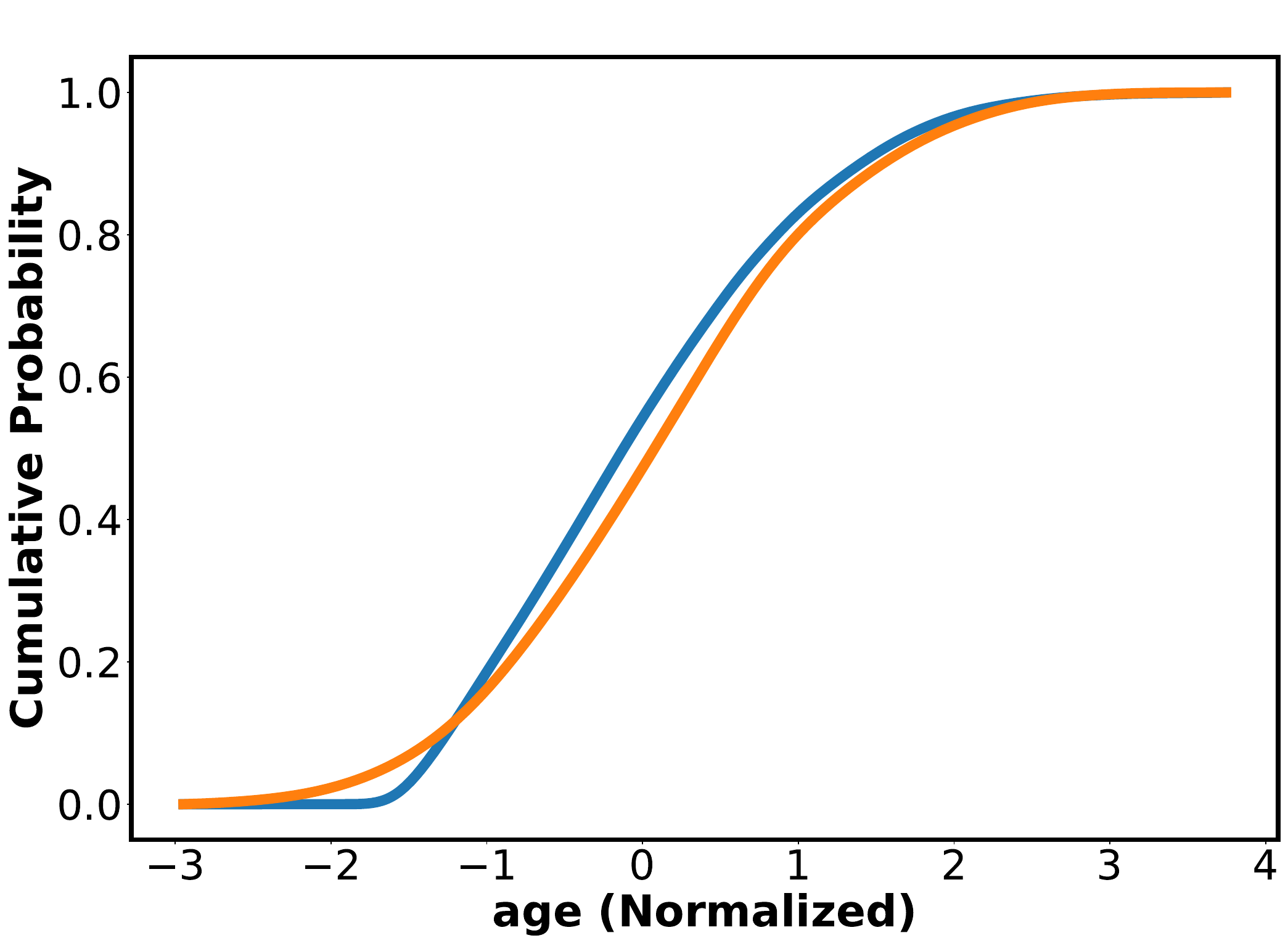}
        \caption{age}
        \label{fig:age}
    \end{subfigure}
    \hfill
    \begin{subfigure}[t]{0.4\linewidth}
        \centering
        \includegraphics[width=\linewidth]{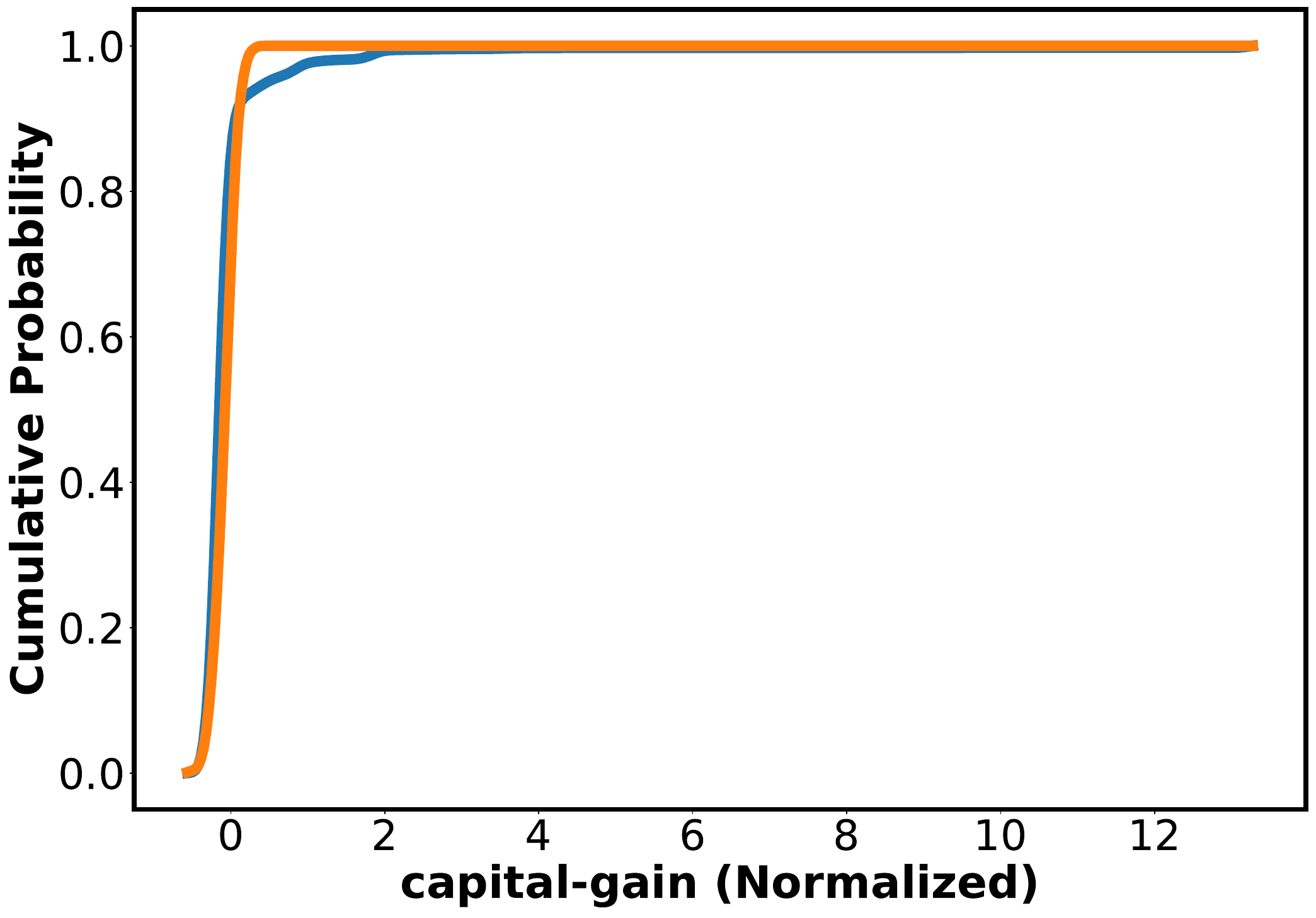}
        \caption{capital-gain}
        \label{fig:capital-gain}
    \end{subfigure}

    \vspace{0.3cm}

    % -------- Second row --------
    \begin{subfigure}[t]{0.45\linewidth}
        \centering
        \includegraphics[width=\linewidth]{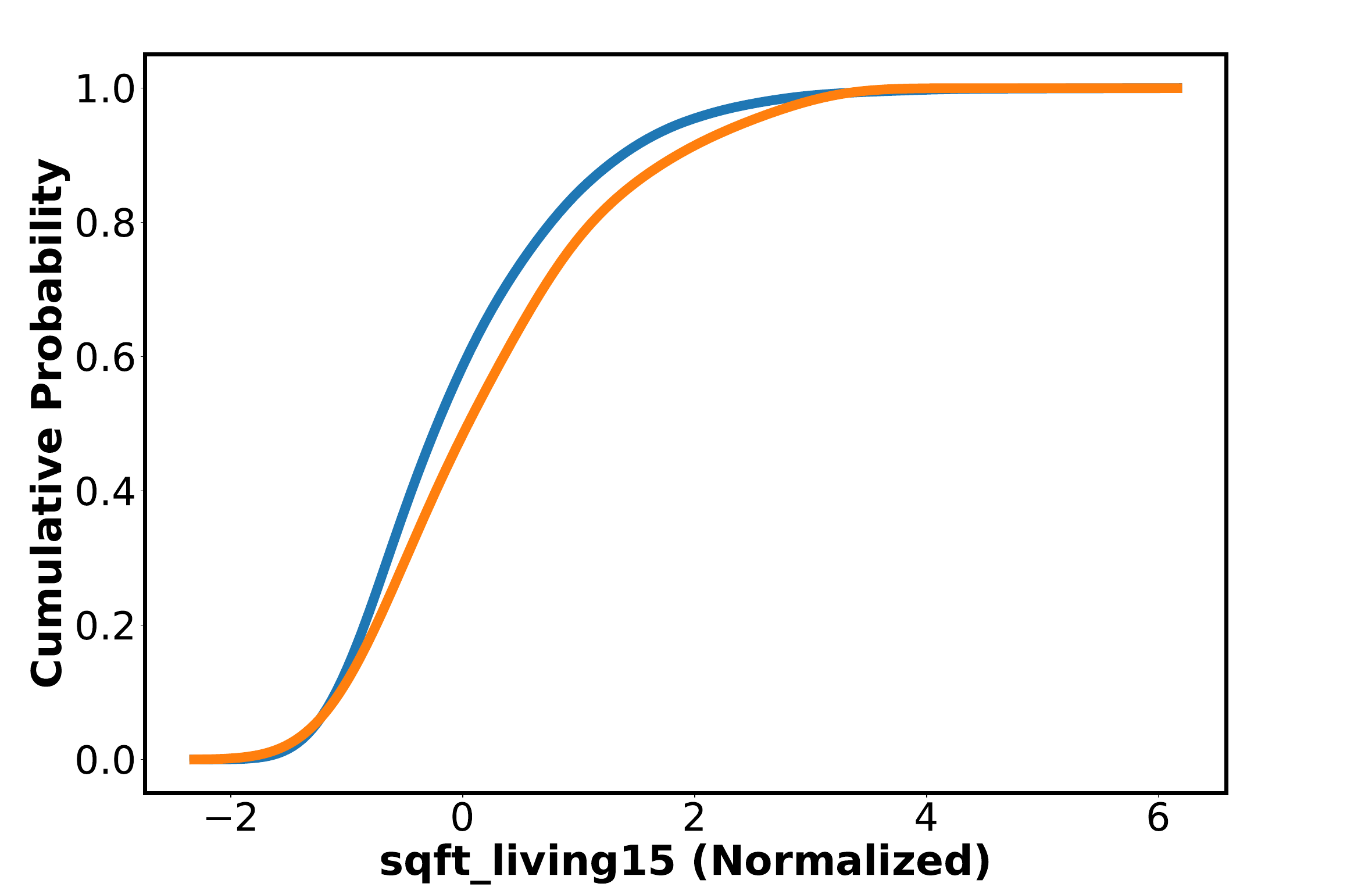}
        \caption{sqft\_living15}
        \label{fig:sqft_living15}
    \end{subfigure}
    \hfill
    \begin{subfigure}[t]{0.4\linewidth}
        \centering
        \includegraphics[width=\linewidth]{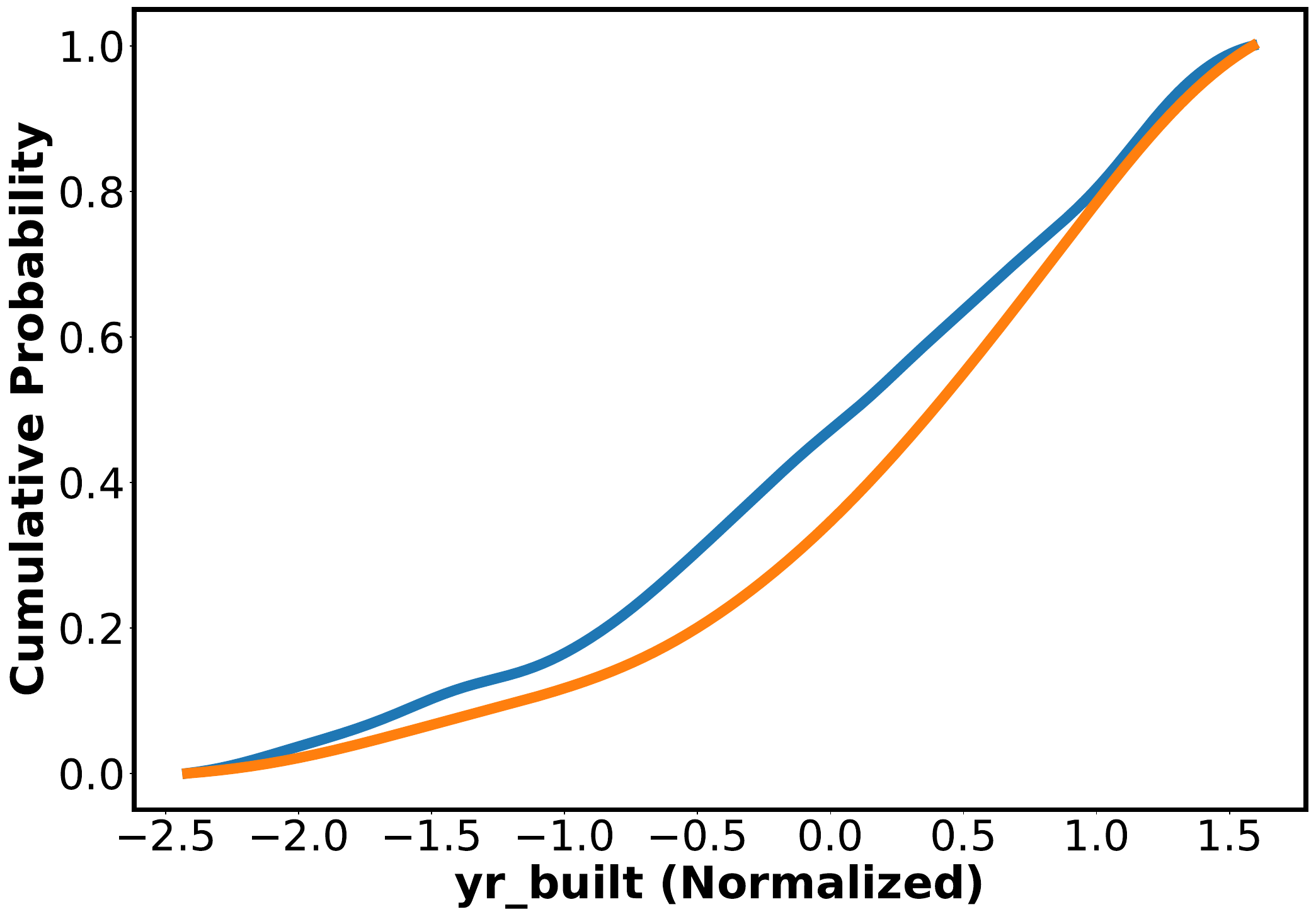}
        \caption{yr\_built}
        \label{fig:yr_built}
    \end{subfigure}

    \caption{CDF visualizations for a subset of features from the Adult and King datasets. The figure includes four CDF plots corresponding to (a) age, (b) capital-gain, (c) sqft\_living15, and (d) yr\_built. The first two plots correspond to the Adult dataset, while the remaining two correspond to the King dataset.}
    \label{fig:adult_and_king_CDF}
\end{figure}

To further examine the distributional alignment between real and synthetic data, we analyze representative features from both a classification dataset and a regression dataset. Specifically, we consider one of the best-matched features and one of the worst-matched features to understand how well the model captures different distribution patterns. The smoothed cumulative distribution function (CDF) curves for these features are shown in Figure~\ref{fig:adult_and_king_CDF}. CDFs provide an intuitive way to characterize data distributions, and smoothing helps reveal overall trends while reducing noise, enabling a clearer assessment of how well the synthetic data reproduces the statistical properties of the real data. 

For the Adult dataset, Sub-figure~\ref{fig:age} and Sub-figure~\ref{fig:capital-gain} show the best and worst CDF plots for age and capital-gain features, respectively. The CDF for the age feature shows a strong overlap between the real and synthetic data, indicating that QTabGAN effectively captures the overall distribution. For capital-gain, small differences are visible in some regions; however, the synthetic CDF closely follows the general cumulative trend of the real data. Such behaviour is expected for highly skewed and heavy-tailed features like capital-gain. This suggests that the model can effectively capture non-uniform and long-tailed distributions without introducing large deviations. Overall, QTabGAN preserves the main distributional characteristics across both smoothly varying and more challenging features, demonstrating stable and robust distributional learning. 

For the King dataset, Sub-figure~\ref{fig:sqft_living15} and Sub-figure~\ref{fig:yr_built} show the best and worst plots for sqft\_living and yr\_built features, respectively. The CDF for sqft\_living15 closely matches the real and synthetic data, indicating that QTabGAN effectively captures the cumulative distribution. For yr\_built, the CDF shows a smooth and monotonic trend, indicating that the proposed model is able to learn the overall distribution of the feature across its full range. The close alignment of the curves suggests that both lower and higher values are adequately represented, reflecting stable learning behaviour. Importantly, the model maintains consistency not only in central regions of the distribution but also toward the tails, which are often more difficult to approximate accurately. These results show that the model preserves the global statistical characteristics in both cases and exhibits consistent performance across features in the King dataset.

\subsection{QTabGAN versus TabularQGAN: A Performance Perspective}
In this section, we compare and contrast our work with TabularQGAN~\cite{bhardwaj2025tabularqgan}, which, to the best of our knowledge, is the only other work on tabular data generation in the quantum computing paradigm. First, we discuss the architectural advantages of QTabGAN that enable it to outperform TabularQGAN. Then, in the result analysis section, we also substantiate our arguments by comparing ML utility and Statistical similarity metrics. We note the following key architectural advantages of QTabGAN.
%The only other work on tabular data generation using quantum computing is TabularQGAN.
\begin{itemize}
    \item TabularQGAN uses a full quantum generator, whereas QTabGAN uses a hybrid quantum-classical generator. In TabularQGAN, the features of the tabular data are encoded through dedicated quantum registers. As a result, the number of qubits required by the model grows linearly with the number of features to be generated. For example, to generate tabular data with 3 to 4 features, TabularQGAN requires 10 to 15 qubits. This increases the circuit size and computation cost.  In contrast, QTabGAN uses a fixed number of qubits to generate a probability distribution over $2^{n}$ outcomes, which is then mapped to multiple tabular features using classical post-processing. This allows QTabGAN to represent and generate a larger number of features without increasing the number of qubits proportionally, making the model more scalable and efficient in terms of quantum resources.

 % QTabGAN fully utilizes  \textcolor{red}{to be filled after discussion with rakesh on relation between features and qubits}. \rak{The number of qubits vs features scaling in Tabular QGAN is linear and in our QtabGAN its exponential that means we can generate exponential number of features with the same number of qubits whereas Tabular GAN could only do it linearly. We are able to do it exponential because we are mapping the 2^n generated prob dists to the features.}
%QTabGAN’s hybrid quantum–classical architecture offers practical advantages over the fully quantum generator used in TabularQGAN \cite{bhardwaj2025tabularqgan}. In TabularQGAN, the number of qubits grows with the number of selected features, since each numerical and categorical feature is encoded through dedicated quantum registers. As a result, even when using only 3-4 features, the model requires 10-15 qubits, which increases circuit size and computational cost. In contrast, QTabGAN employs a fixed-size quantum circuit, making the model lighter, more stable, and easier to scale.

    \item TabularQGAN represents numerical features using discretized qubit registers. This encoding approach divides continuous values into discrete quantum states, which restricts the resolution with which numerical ranges can be represented. QTabGAN avoids such binning by mapping features directly to the continuous interval $[0, \pi]$, allowing the quantum circuit and classical mapper to operate on smooth, full-resolution inputs.

    \item TabularQGAN has limited feature coverage due to the qubit overhead for feature-specific encoding. TabularQGAN experiments are limited to a small subset of features. As a result, only part of the available quantum state space can be explored. In contrast, QTabGAN generates a full $256$-dimensional probability vector from its 8-qubit circuit, and its classical network learns relationships across all features in the dataset. This avoids feature-selection constraints and enables the model to capture broader patterns.

    \item QTabGAN’s hybrid structure makes it more suitable for present-day quantum hardware. By using fewer qubits and relying on classical neural networks for scalable learning, QTabGAN avoids the heavy qubit requirements and deeper circuits approach of TabularQGAN, reducing sensitivity to noise and making the model more practical for real-world synthetic tabular data generation.
\end{itemize}

\subsubsection{NISQ Compatibility and Efficiency}
QTabGAN is highly efficient for Noisy Intermediate-Scale Quantum (NISQ) devices. Specifically, QTabGAN is designed to operate with fewer qubits, fewer quantum gates, and a shallow circuit depth, factors that are critical for practical deployment on current quantum hardware. By employing a circular (ring) entanglement structure, the model leverages nearest-neighbour entangling operations, which have been shown to provide sufficient expressivity while limiting circuit depth and reducing noise accumulation in variational quantum circuits\cite{benedetti2019parameterized}. This architecture effectively reduces common NISQ bottlenecks such as decoherence and gate errors, thereby enhancing computational stability and fidelity. Consequently, QTabGAN not only achieves high-quality data generation performance but also maintains quantum resource efficiency, making it well-suited for implementation on near-term quantum devices. Due to limited access to quantum hardware and the high operational costs associated with current NISQ devices, all experiments were conducted on standard simulators, a widely adopted practice in contemporary quantum machine learning research.

%% file: conclusion.tex
\section{Conclusion}
\label{VII}

This study demonstrates the effectiveness of our QTabGAN model for tabular data, which integrates a quantum circuit with classical neural networks to generate high-quality synthetic data for both classification and regression datasets. The proposed model not only leverages the expressive power of quantum circuits to capture intricate feature interactions in high-dimensional tabular datasets, but also utilizes classical deep learning components to ensure scalability and stability during training. The model showed strong predictive performance and favourable statistical similarity differences across diverse datasets. In our experiments, QTabGAN outperformed several classical generative baselines as well as the quantum baseline, demonstrating superior capability in producing synthetic data that closely matches real data. By embedding quantum circuits into the data generation process, the model effectively captures non-linear dependencies and complex probability distributions that are often difficult for classical models to represent. This property makes QTabGAN particularly valuable in scenarios where synthetic data must maintain a realistic structure. Overall, the results suggest that quantum-enhanced generative models can offer significant benefits in domains requiring secure and high-fidelity data synthesis, such as finance, healthcare, and IoT security. The ability of QGANs to balance utility along with generating realistic tabular data highlights their potential as a promising next-generation approach to synthetic data generation.

%% file: limitations.tex
\section{Limitations and Future Scope}
\label{Limit}

Our model faces limitations due to the computationally expensive training of quantum circuits. As quantum hardware continues to grow in size and fidelity, performing training and sampling at scale on real quantum hardware would be valuable for understanding the impact of device noise on sample quality. Future work could explore more balanced hybrid configurations, as the incorporation of quantum components in the generator naturally increases its expressive capacity relative to the classical discriminator. Exploring architectures that combine these abilities may further improve training stability and generative performance.